\documentclass{article}

\usepackage[utf8]{inputenc}
\usepackage[T1]{fontenc}
\usepackage{wrapfig}
\usepackage{arxiv}
\usepackage{subcaption}
\usepackage{mathtools}
\usepackage{setspace}
\usepackage{graphicx}
\usepackage{float}
\usepackage{csquotes}
\usepackage{placeins}
\usepackage{array,booktabs}
\usepackage{hyperref}
\usepackage{url}
\usepackage{amsfonts}
\usepackage{nicefrac}
\usepackage{microtype}
\usepackage{lipsum}

\graphicspath{{./images/}}

\title{A Physics-Informed U-Net-LSTM Network for Nonlinear Structural Response under Seismic Excitation}

\author{
Sutirtha Biswas\\
Department of Civil Engineering\\
Indian Institute of Technology (BHU) Varanasi\\
Varanasi, Uttar Pradesh, India\\
ORCID: 0009-0006-6411-7189\\
\texttt{sutirtha.biswas.cd.civ21@iitbhu.ac.in}
\And
Kshitij Kumar Yadav\\
Department of Civil Engineering\\
Indian Institute of Technology (BHU) Varanasi\\
Varanasi, Uttar Pradesh, India\\
ORCID: 0000-0001-5287-6635\\
\texttt{kshitij.civ@iitbhu.ac.in}
}

\begin{document}
\maketitle
\begin{abstract}
Accurate and efficient seismic response prediction is essential for the design of resilient structures. While the Finite Element Method (FEM) remains the standard for nonlinear seismic analysis, its high computational demands limit its scalability and real-time applicability. Recent developments in deep learning—particularly Convolutional Neural Networks (CNNs), Recurrent Neural Networks (RNNs), and Long Short-Term Memory (LSTM) models have shown promise in reducing the computational cost of the nonlinear seismic analysis of structures. However, these data-driven models often struggle to generalize and capture the underlying physics, leading to reduced reliability. We propose a novel Physics-Informed U-Net-LSTM framework that integrates physical laws with deep learning to enhance both accuracy and efficiency. The proposed 1D U-net captures the underlying latent features of the long-term input sequences. And by embedding domain-specific constraints into the learning process, the proposed model achieves improved predictive performance over conventional Machine Learning (ML) architectures. This approach bridges the gap between purely data-driven methods and physics-based modeling, offering a robust and computationally efficient alternative for predicting the seismic response of structures. 
\end{abstract}


\section{Introduction}
The increasing number and intensity of natural disasters such as earthquakes, typhoons, and tsunamis have heightened the global demand for resilient and adaptive infrastructure systems. Among these hazards, seismic events pose particularly severe threats to structural integrity, often resulting in widespread damage and significant loss of life. To ensure structural safety and reliability, and to limit potential damage, accurate prediction of structural response under seismic loading is essential. Conventional numerical methods, such as the Newmark-$\beta$ method \cite{newmark1959method}, KR-$\alpha$ \cite{kolay2014development}, and the finite element method (FEM)\cite{reddy1993introduction,huebner2001finite,zienkiewicz2005finite}, have long been employed to model the dynamic behavior of structures. These techniques are reliable and firmly grounded in physical principles; however, they are computationally intensive, particularly when applied to nonlinear analysis of full-scale structures. The high computational cost associated with these methods limits their suitability for real-time applications and large-scale parametric studies. 

In recent years, machine learning (ML) and artificial intelligence (AI) have emerged as promising alternatives or supplements to traditional numerical methods. A range of ML techniques, including support vector machines (SVM) \cite{yinfeng2008nonlinear,segura2020metamodel,gharehbaghi2020estimating}, multilayer perceptrons (MLP) \cite{NN_ref1,NN_ref2,NN_ref3,NN_ref4,NN_ref5,NN_ref6,NN_ref7,NN_ref8,NN_ref9,lagaros2012neural}, deep neural networks \cite{vaidyanathan2005artificial,kim2019response,kim2020probabilistic}, long short-term memory networks (LSTMs) \cite{zhang2019deep}, and convolutional neural networks (CNNs) \cite{wu2019deep,lecun1995convolutional,lecun1989handwritten,lecun2002gradient,krizhevsky2012imagenet}, have been employed to develop surrogate models for the prediction of the seismic response of structures. These approaches reduce computational cost while offering acceptable levels of accuracy. RNNs and LSTMs, in particular, are often adopted for modeling temporal dynamics in structural systems due to their ability to learn long-term dependencies in time series data \cite{graves2012long}. Whereas, CNNs with dilated filters have shown promise for time-series modeling by enabling access to long-range dependencies without recurrence \cite{bai2018empirical,oord2016wavenet}. These models have also been successfully applied in domains such as ECG classification, structural health monitoring, continuum damage mechanics and anomaly detection \cite{bird2020cross,tang2019convolutional,danilczyk2021smart,zhao2019speech,pantidis2023integrated,pantidis2024fenn}.

Inspired by these advances, a new class of ML models, physics-informed machine learning, has gained traction within the earthquake engineering and structural dynamics community, where it addresses critical challenges such as data scarcity and limited model generalization. By embedding governing physical laws directly into ML architectures, physics-informed frameworks enhance predictive accuracy and ensure physical consistency. For instance, physics-informed LSTM frameworks (e.g., PhyLSTM2 and PhyLSTM3) \cite{phylstm} estimate key variables such as displacement and restoring force by incorporating these laws into the loss function during training. Similarly, PhyCNN \cite{phycnn} integrates domain knowledge with deep CNNs to predict the nonlinear seismic response of structures through a data-driven approach. These innovative methods have been applied across diverse scenarios, including subway station dynamics \cite{huang2021deep}, vehicle–bridge interactions \cite{li2021dynamic}, and dam behavior prediction \cite{li2022new}. More recently, PI-LSTM has been introduced and benchmarked against PhyCNN and PhyLSTM, further demonstrating the potential of physics-guided ML architectures for structural response prediction \cite{liu2023pi}. Other works \cite{raissi2018deep,raissi2019physics,sun2020surrogate,zhu2019physics} have highlighted the importance of incorporating physical constraints—such as ordinary and partial differential equations and boundary conditions—directly into deep learning models to improve stability and accuracy, even when data are limited. 

Despite this progress, key limitations persist. LSTM and RNN models, while effective in modeling temporal dependencies, suffer from long training times and are prone to overfitting, particularly in low-data regimes \cite{bengio1994learning,trinh2018learning}. Deep CNN models, although powerful, require substantial computational resources when applied to long time series, and their performance degrades under large plastic deformations \cite{kuo2024gnn}. Moreover, capturing long-range interactions and nonlinearities in dynamic systems remains a challenge. 
To overcome these limitations and address the growing demand for reliable seismic response prediction, we propose a novel Physics-Informed U-Net–LSTM (PhyULSTM) framework. While U-Net architectures have been predominantly employed in image processing tasks such as medical imaging and semantic segmentation, we reformulate the U-Net into a causal one-dimensional architecture specifically designed for time-series modeling of structural response and ground motion signals.

The proposed 1D U-Net extracts hierarchical, multi-scale temporal representations while strictly preserving chronological causality—an essential requirement for structural health monitoring and early-warning applications. By leveraging the slow feature hypothesis \cite{wiskott2002slow}, the encoder–decoder structure captures temporally coherent latent features that evolve smoothly across scales, enabling robust representation of nonlinear structural dynamics.
These learned representations are subsequently processed through a deep LSTM network, which is excellent for modeling long-range temporal dependencies and complex hysteretic behavior inherent in nonlinear structural systems. Beyond purely data-driven training, physical knowledge in the form of governing equations of motion is embedded directly into the loss function via a physics-based residual term. This ensures mechanical consistency and enables accurate prediction even in data-scarce or partially observed scenarios. 
The framework is validated through both numerical and experimental benchmarks, where it consistently and significantly outperforms popular physics-informed models such as PhyLSTM and PhyCNN. Its superiority is firmly established through rigorous quantitative assessment, including high correlation coefficients between predicted and actual responses and superior global error metrics. Furthermore, detailed analysis of hysteresis plots, Fourier amplitude spectra, and peak response errors reveals that the model captures the system's nonlinear behavior with exceptional fidelity, accurately reproduces critical frequency components, and precisely predicts peak responses—capabilities that are essential for subsequent fragility and reliability analysis of structures.

\section{Physics-Informed U-net-LSTM Network}
\label{sec:PhyULSTM}

\begin{figure*}
\centering
\includegraphics[width=\textwidth]{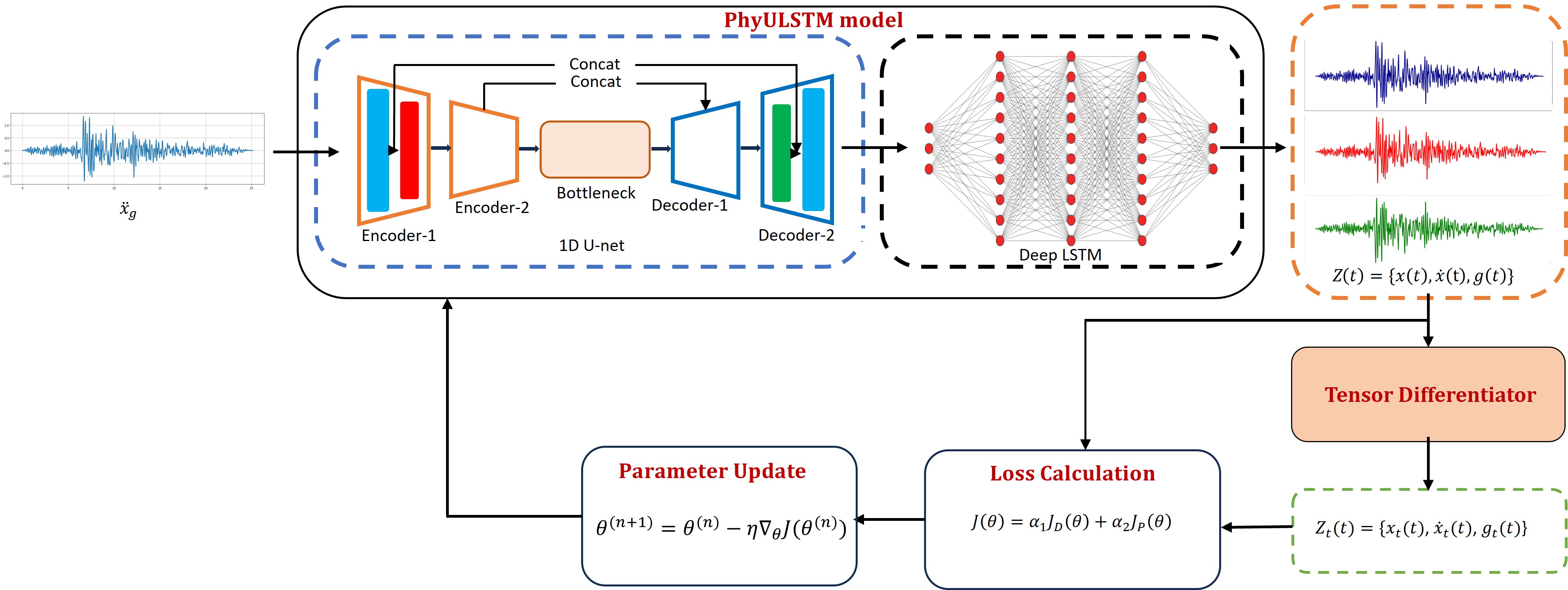} 

\caption{The proposed Physics-Informed U-net-LSTM Network (PhyULSTM) for time-series modeling. The 1D U-net first receives the ground acceleration data as input and extracts features at
multiple time scales. The outputs from the U-net are then fed to the deep LSTM network that maps the temporal feature maps to the corresponding output space. The outputs are state space variables 
${z}(t)$ = $\{x (t), \dot{x}(t), g(t)\}$.
Available physics knowledge is incorporated directly into the loss function. In addition to it, a tensor differentiator is implemented to calculate the derivative of state space outputs ${z_t}(t)$ = $\{x_{t} (t), \dot{x}_{t}(t), g_t (t)\}$ to construct the physics loss from the governing equation. By optimising the network hyperparameters, PhyULSTM can interpret the measurement data (e.g. $\{x_m, \dot{x}_m, g_m\}$) while satisfying the physical equation of motion in equation \ref{eq:gov}, e.g. $f\rightarrow 0$.}
\label{Fig1}
\end{figure*}

In this section, we present the PhyULSTM framework for surrogate modeling of nonlinear dynamic structural systems subjected to seismic ground motion. PhyULSTM consists of three core components: a 1D U-Net, a deep LSTM network, and a tensor differentiator. The proposed 1D U-Net, a tailored variant of the standard U-Net architecture, is designed to work synergistically with deep LSTM networks for time-series analysis. It has potential to alleviate common limitations of traditional RNNs and LSTMs—such as difficulty in capturing long-range temporal dependencies and susceptibility to overfitting—especially when processing high-resolution seismic inputs like ground acceleration. This is achieved by leveraging the ‘Slow Feature’ hypothesis \cite{wiskott2002slow}, which posits that the most informative features in temporal sequences evolve slowly over time, even when raw signals exhibit rapid fluctuations. By extracting such temporally stable features across multiple time scales and applying causal convolutions—which restrict each prediction to depend only on current and past inputs—the U-Net enhances the LSTM’s ability to learn long-term dependencies within an auto-regressive framework. Additionally, physical knowledge is explicitly incorporated into the model through a tensor differentiator (based on the finite difference method), as introduced by Zhang et al. \cite{phycnn}. The differentiator component computes derivatives of the state-space outputs to construct a physics-informed loss term derived from the system’s governing equations. The hybrid U-Net–LSTM model is trained on seismic input–output datasets—obtained either through numerical simulation or sensor measurements—and subsequently deployed as a surrogate model for predicting structural responses under future seismic excitations. A detailed illustration of the  PhyULSTM framework is presented in Figure \ref{Fig1}.

To demonstrate the PhyULSTM framework, we consider a general dynamic structural system subjected to ground motion excitation, governed by the equation:

\begin{equation}
M \ddot{x}(t) + h(t) = -M \Gamma \ddot{x}_g(t),
\label{eq:New2}
\end{equation}

where \(M\) is the mass matrix; \(x(t)\) and \(\dot{x}(t)\) denote the relative displacement and velocity with respect to the ground; \(\ddot{x}_g(t)\) is the ground acceleration vector; \(\Gamma\) represents the force distribution vector; and \( h(t) \) denotes the latent generalized restoring force vector, which is modeled as a function of the ground acceleration \( \ddot{x}_g(t) \) and the acceleration of structures with respect to the ground \( \ddot{x}(t) \). Normalizing Eq. \eqref{eq:New2} by \(M\), the governing equation can be rewritten as

\begin{equation}
f : \ddot{x}(t) + g(t) + \Gamma \ddot{x}_g(t) \rightarrow 0,
\label{eq:gov}
\end{equation}

where \(g(t) = M^{-1} h(t)\) is the mass-normalized restoring force. 
\subsection{1D U-net}
\label{subsec:Proposed 1D U-net:}
\begin{figure*}
\centering
\includegraphics[height=3.3in]{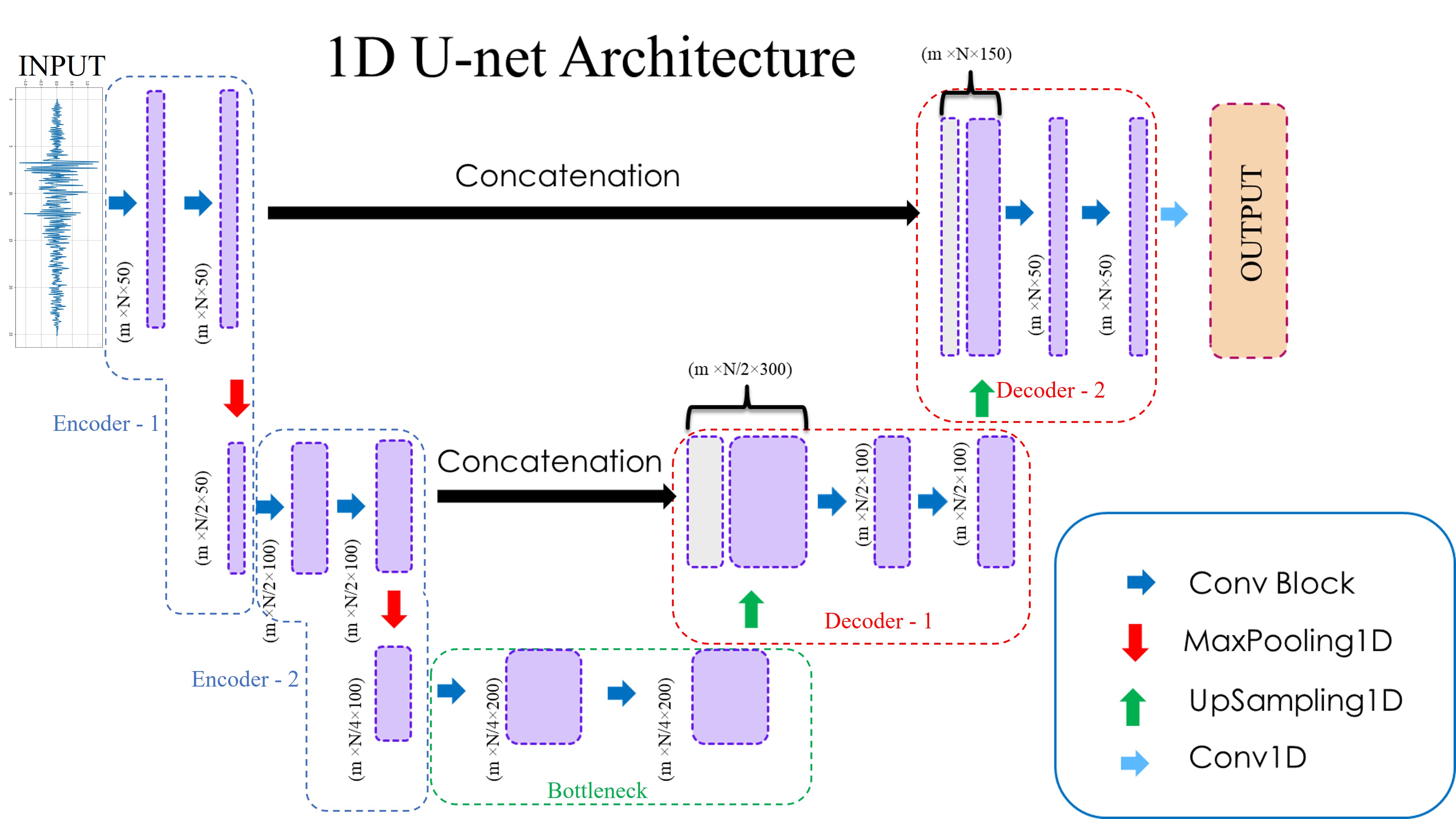} 
\caption{Proposed 1D version of U-net: The U-Net architecture is mainly divided into four major components: the encoder blocks, bottleneck, decoder blocks, and the output convolution block. The encoder blocks perform convolution and pooling operations, generating skip connections and providing input to subsequent encoder blocks after pooling. The decoder blocks involve upsampling, followed by concatenation with the skip connections from the corresponding encoder block outputs, and subsequent convolution operations. The bottleneck layer, serving as a bridge, consists solely of a convolution block, linking the encoder and decoder paths. The output from the final decoder block is fed to the output convolution block, where it first passes through a Conv1D layer with a number of filters equal to the number of channels required in the final output, and sigmoid activation, followed by a second Conv1D layer and a linear activation function, which further processes the output to format the final output as a three-dimensional array, where the entries are sampled in the first dimension, time history steps in the second dimension, and output features in the last dimension. Here, \textquoteleft ConvBlock\textquoteright is a fundamental building block consisting of a convolutional layer, followed by Batch Normalisation, and a ReLU activation function.}

\label{U-net_arch}
\end{figure*}

U-Net \cite{ronneberger2015u} is an established framework originally developed for biomedical image segmentation \cite{unet1,unet2}. It has a symmetrical U-shaped structure, consisting of an encoder and a decoder connected by skip connections. The encoder captures contextual information by progressively downsampling the input image, while the decoder reconstructs the image by upsampling and combining high-resolution features from the encoder, ensuring precise localisation.
The ‘Slow Feature Analysis’ hypothesis \cite{wiskott2002slow} posits that meaningful features in dynamic signals often evolve slowly over time. Considering the encoder-decoder components and the ability to capture contextual information and combine resolution features, U-nets compute features at different time scales with two-dimensional convolutions and combining them to make predictions at the same resolution as the input. One such adaptation is the one-dimensional U-Net, suitable for time-series analysis \cite{stoller2019seq}. In this study, we propose a causal 1D U-Net, as shown in Figure \ref{U-net_arch} and represented in the table \ref{tab:unet_architecture},   to process seismic time-series data that captures such features across multiple temporal scales while maintaining causal structure, ensuring outputs at a given time step depend only on current and past inputs. This enhances the model's suitability for sequential tasks and integration with LSTM networks.

The architecture comprises four parts: the encoder (contractive path), the bottleneck, the decoder part (expansive path) and finally, the output convolution block.
The input is processed sequentially through two encoder blocks, having 50 and 100 filters respectively. Each encoder block generates two outputs: one is a skip connection (or shortcut connection) for the respective decoder block, and the other (output after 1D pooling operation) serves as the input for the subsequent encoder block. The final encoder block provides inputs for the bottleneck layer and also creates a skip connection.
The encoder block comprises two sequential ‘ConvBlock' units. Each ‘ConvBlock' executes a series of operations: a 1D convolution ( kernel size = 2, padding =  ‘causal’ ), followed by Batch Normalization, and finally a ReLU activation function. The resulting output from this process serves as both a skip connection (shortcut path) and as the input for the subsequent MaxPooling1D operation. MaxPooling1D is a down-sampling operation typically used in convolutional neural networks (CNNs). This method reduces the dimensionality of the input tensor by extracting the maximum value from within a sliding window, known as the pool size (set to 2), across the input. So, the encoder block consists of two main components: the convolution blocks and the max pooling block.

For down-sampling, the ‘Conv1D’ operation with strides=2 (padding= ‘causal’) can also be implemented, but here in this case ‘MaxPooling1D’ is used to emphasize keeping feature extraction and down-sampling as distinct operations. And to ensure that the most significant features (e.g. sudden large significant peaks in ground acceleration data) are preserved during downsampling. There are also a few additional advantages. It is more straightforward than implementing ‘Conv1D’ with a stride of 2, and it reduces feature map size without adding extra learnable parameters. The bottleneck layer also comprises two sequential ‘ConvBlock' units, using 200 filters without any pooling operation. This layer acts as a crucial bridge between the encoder and decoder, capturing complex, high-level features while maintaining the spatial dimensions of the feature maps.
In the decoder stage, the first decoder block receives the output from the bottleneck layer, along with the skip connection from the second encoder block, and uses these inputs to upsample the feature map. Within the decoder block, a 1D upsampling operation is performed on the input tensor that repeats each temporal step size times along the time axis. In this case, as the parameter size=2 is specified, the layer performs upsampling by repeating each temporal step of the input sequence twice along the time axis. This step helps to gradually reconstruct the original input size. After the 1D upsampling operation, the feature map is concatenated with the corresponding skip connection from the encoder blocks. Then the output passes through two ConvBlocks. This sequence is repeated in each of the two decoder blocks, with 100 and 50 filters, respectively, to progressively refine the feature maps and reconstruct the final output with high accuracy and detail.

The output from the final decoder block is then first passed through a Conv1D layer with a number of filters equal to the number of channels required in the final output, a kernel size of 1, and sigmoid activation, which reduces the channels to match the dimension of the final output required and maps the output values between 0 and 1. The output has the same length as the input due to ‘same' padding. This is followed by a second Conv1D layer and a linear activation function, which further processes the output by expanding the channels to format the final output as a three-dimensional array, where the entries are sampled in the first dimension, time history steps in the second dimension, and output features in the last dimension, this is very crucial otherwise the deep LSTM network cannot be implemented subsequently.
The network architecture is shown in Figure \ref{U-net_arch} and represented in the table \ref{tab:unet_architecture}. The U-Net output feeds directly into the LSTM module.
\begin{table*}[h!]
\centering
\caption{Proposed 1D U-Net Architecture}
\label{tab:unet_architecture}
\begin{tabular}{l c c c c c}
\toprule
\textbf{Layer} & \textbf{Operation} & \textbf{Filters} & \textbf{Kernel} & \textbf{Stride} & \textbf{Activation} \\
\midrule
Encoder 1 & Conv Block (x2)  & 50  & 2 & 1 & ReLU \\
Encoder 1 & Max Pooling            & --  & 2 & 2 & --   \\

Encoder 2 & Conv Block (x2)  & 100 & 2 & 1 & ReLU \\
Encoder 2 & Max Pooling            & --  & 2 & 2 & --   \\

Bottleneck & Conv Block (x2) & 200 & 2 & 1 & ReLU \\

Decoder 1 & UpSampling1D           & --  & -- & -- & --   \\
Decoder 1 & Conv Block (x2)  & 100 & 2 & 1 & ReLU \\

Decoder 2 & UpSampling1D           & --  & -- & -- & --   \\
Decoder 2 & Conv Block (x2)  & 50  & 2 & 1 & ReLU \\

Output & Convolutional & $N_{\text{out}}$ & 1 & 1 & Sigmoid \\
Final Output & Convolutional & $N_{\text{out}}$ & 1 & 1 & Linear \\
\bottomrule
\end{tabular}

\vspace{2mm}
\footnotesize
\textit{Note: Each Conv Block consists of consecutive operations: Convolution $\rightarrow$ Batch Normalization $\rightarrow$ ReLU activation.}
\end{table*}

\subsection{Deep LSTM Network}
\label{subsec:Deep LSTM}

\begin{figure*}
\centering
\begin{subfigure}[b]{\textwidth}
    \centering
    \includegraphics[width=0.85\textwidth, height=3in]{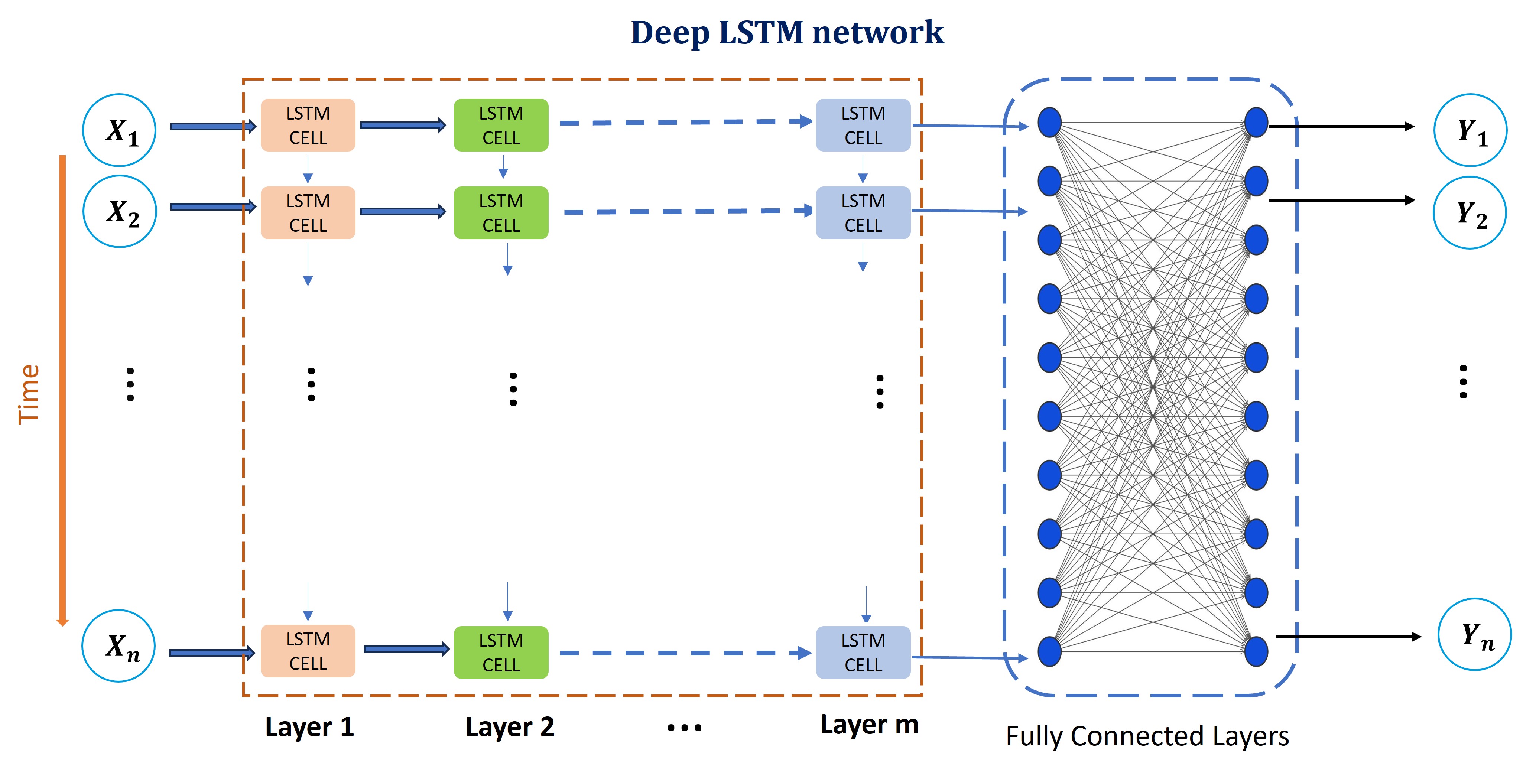} 
    \caption{}
    \label{fig:LSTM}
\end{subfigure}

\begin{subfigure}[b]{0.85\textwidth} 
    \centering
    \includegraphics[width=0.7\textwidth, height=2.5in]{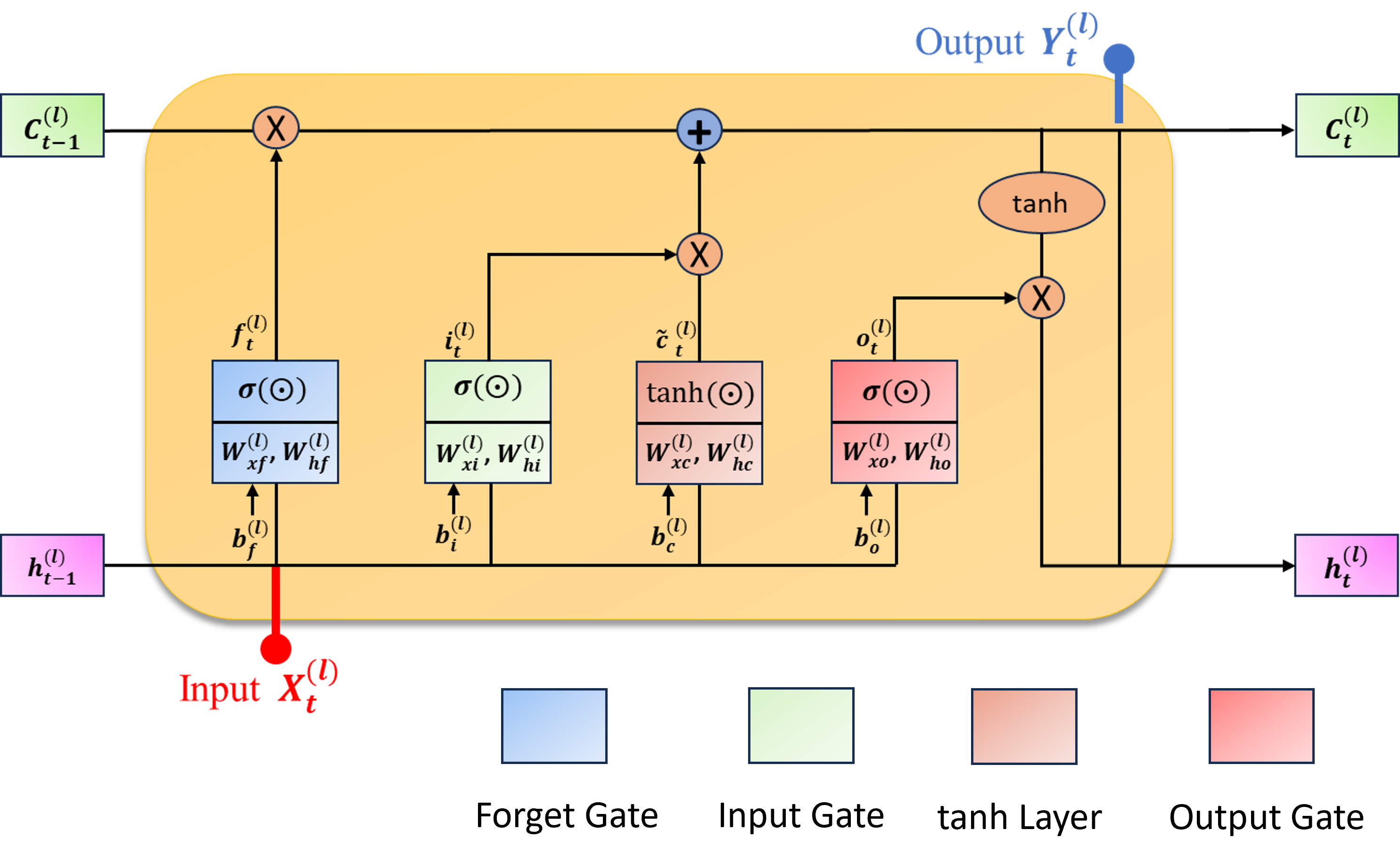} 
    \caption{}
    \label{fig:LSTM-Cell}
\end{subfigure}

\caption{Schematic of deep LSTM networks: (a)  corresponds to the network architecture of a deep LSTM network featuring \(m\) LSTM layers and multiple fully-connected layers for sequence-to-sequence modeling and (b) corresponds to the architecture of a typical LSTM cell at the \(l\)th layer and time \(t\), showing cell input \(X^{(l)}_t\), cell output \(Y^{(l)}_t\), cell state \(c^{(l)}_t\), hidden state \(h^{(l)}_t\), and gate variables \(\{f^{(l)}_t, i^{(l)}_t, \tilde{c}^{(l)}_t, o^{(l)}_t\}\), respectively.}
\label{fig:LSTM-overall}
\end{figure*}

Each LSTM layer consists of a set of LSTM cells, as depicted in Figure \ref{fig:LSTM-overall}. Like the neural nodes in traditional ANNs, each LSTM cell contains an independent set of weights and biases shared across the entire temporal sequence within the layer. First, the basic architecture of the deep LSTM network for sequence-to-sequence modeling was presented by \cite{zhang2019deep}. The architecture includes a number of hidden layers - additional to input and output layers - as shown in Figure \ref{fig:LSTM-overall}(a), some of which are LSTM layers, while others are fully connected layers. The LSTM cell includes four components: an internal cell, an input gate, a forget gate, and an output gate.
The internal cell remembers the cell state from the previous time step via a self-recurrent connection. The input gate maintains the flow of input activations into the internal cell state, while the output gate controls the flow of output activations from the LSTM cell. The forget gate scales the internal cell state, allowing the LSTM cell to adaptively forget or reset its memory. Through the coordinated actions of the input, forget, and output gates, the cell state can selectively transmit essential information along the temporal sequence, capturing both long-term and short-term dependencies in a dynamic system effectively.

Let us consider at time step t (where ( $t = 1, 2, 3, \ldots, n $ ); with n being the total number of time steps) and within the   $l$-th LSTM network layer, the input state to the LSTM cell as $x_t^{(l)}$, and the forget gate as $f_t^{(l)}$, the input gate as  $i_t^{(l)}$, the output gate as $o_t^{(l)}$, the cell state memory as $c_t^{(l)}$ and hidden state memory as $h_t^{(l)}$.
Similarly, for the previous time step t-1, we denote the cell state memory as  $ c_{(t-1)}^{(l)} $ and hidden state memory as $ h_{(t-1)}^{(l)} $. The relationship between these variables can be expressed as follows:
\begin{equation}
f_t^{(l)} = \sigma \left( W_{xf}^{(l)} \cdot x_t + W_{hf}^{(l)} \cdot h_{t-1}^{(l)} + b_f^{(l)} \right)
\label{eq:LSTM1}
\end{equation}
\begin{equation}
i_t^{(l)} = \sigma \left( W_{xi}^{(l)} \cdot x_t + W_{hi}^{(l)} \cdot h_{t-1}^{(l)} + b_i^{(l)} \right)
\label{eq:LSTM2}
\end{equation}
\begin{equation}
\tilde{c}_t^{(l)} = \tanh \left( W_{xc}^{(l)} \cdot x_t + W_{hc}^{(l)} \cdot h_{t-1}^{(l)} + b_c^{(l)} \right)
\label{eq:LSTM3}
\end{equation}
\begin{equation}
o_t^{(l)} = \sigma \left( W_{xo}^{(l)} \cdot x_t + W_{ho}^{(l)} \cdot h_{t-1}^{(l)} + b_o^{(l)} \right)
\label{eq:LSTM4}
\end{equation}
\begin{equation}
c_t^{(l)} = f_t^{(l)} \odot c_{t-1}^{(l)} + i_t^{(l)} \odot \tilde{c}_t^{(l)} 
\label{eq:LSTM5}
\end{equation}
\begin{equation}
h_t^{(l)} = o_t^{(l)} \odot \tanh \left( c_t^{(l)} \right)
\label{eq:LSTM6}
\end{equation}
Where $W_{(\alpha \beta)}^{(l)} $ (with $\alpha$ = ${\{x,h\}}$  and $\beta= {\{f,i,c,o\}}$ ) denotes weight matrix corresponding to different inputs (e.g. $x_t^l$  or  $h_t^l$ ) different gates  (e.g., input gate, forget gate, tanh layer, or output gate). And  $b_\beta^{(l)}$ represents the corresponding bias vectors; the lth layer of the LSTM network is denoted by the superscript l. Here, $\tilde{c}_t^{(l)}$ denotes a vector of intermediate candidate values created by a tanh layer shown in Figure\ref{fig:LSTM-Cell}; $\sigma$ is the logistic sigmoid function; tanh is the hyperbolic tangent function; $\odot$ denotes the Hadamard product (elementwise product). The complex connection system within each LSTM cell makes the deep LSTM network powerful in sequence modelling, in which the fully connected layers are beneficial to map the temporal feature maps to the corresponding output space.

The output of the final dense layer produces a time-series prediction of the state-space variables $z(t) = {x(t), \dot{x}(t), g(t)}$, where $x$ is displacement, $\dot{x}$ is velocity, and $g$ is restoring force. While this formulation accurately captures temporal patterns, it lacks explicit knowledge of the governing physical laws.To address this limitation, the predicted outputs are further processed using a tensor differentiator to compute their temporal derivatives. A specialized loss function is then formulated, tailored to the specific requirements of each scenario. The tensor differentiator and the loss function are described in detail in Sections~\ref{subsec: Tensor Differentiator}
and
~\ref{subsec:Loss_function:}, respectively.
Training proceeds by minimizing the loss:
\begin{equation}
\hat{\theta} \coloneqq \arg \min_{\theta} J(\theta)
\label{eq:opt}
\end{equation}

The model is implemented in Python using Keras \cite{chollet2018keras} with TensorFlow backend. Model training is carried out using GPU acceleration. The detailed network architecture of the proposed framework is represented in Figure \ref{Fig1}.

\subsection{Tensor Differentiator}
\label{subsec: Tensor Differentiator}
Numerical differentiation is an essential method widely used in computational mathematics, particularly in solving differential equations and gradient-based optimization. Finite differences (FD) approximate derivatives by combining nearby function values using a set of weights. Here in this study to calculate the derivatives of the state space outputs of the neural network, a tensor differentiator  utilizing finite difference approach is implemented.\cite{phycnn} This section gives detailed explanation about the construction of finite difference matrix ($[\Phi_t]$) to calculate the first order derivatives providing second order accuracy meaning the truncation error is of order 2 (i.e. $O(\Delta t^2)$).
For all the interior points central difference stencil is utilized whereas for the first and last point forward and backward difference stencil is implemented.

Let us assume a function $u = [u_0, u_1, \ldots, u_{n-1}]$ is sampled at time $t_0, t_1, \ldots, t_{n-1}$. Here the data $u_i = u(t_i)$ given at equispaced points $t_i = t_0 + i \Delta t$, for $i = 0, 1, 2, \ldots, n-1$ where $\Delta t$ is the step size. The first order derivatives are approximated as:

\begin{equation} \label{eq:first_derivative_approx}
\left. \frac{du}{dt} \right|_{t_i} \approx
\begin{cases}
\displaystyle \frac{-\frac{3}{2}u_i + 2u_{i+1} - \frac{1}{2}u_{i+2}}{\Delta t}, & i = 0 \\
\displaystyle \frac{u_{i+1} - u_{i-1}}{2\Delta t}, & 0 < i < n-1 \\
\displaystyle \frac{\frac{1}{2}u_{i-2} - 2u_{i-1} + \frac{3}{2}u_i}{\Delta t}, & i = n-1
\end{cases}
\end{equation}
In order to compute the first-order derivative of vector $\{u\} = [u_0, u_1, \ldots, u_{n-1}]^T$ as mentioned in equation \ref{eq:first_derivative_approx}, which is $\{u_t\}$ ,  matrix $[\Phi_t]$ is constructed which transforms $\{u\}$ to $\{u_{t}\}$.
\begin{equation} \label{eq:ut_phi_u}
\{u_t\} = [\Phi_t] \{u\}
\end{equation}

Where $[\Phi_t]$ is an $n \times n$ finite difference matrix whose entries constitute the optimal weights for approximating first derivatives at each discrete point through a weighted combination of neighboring function values.
The complete finite difference matrix can be expressed as:
\[
\Phi_t = \frac{1}{\Delta t}
\begin{bmatrix}
-3/2 & 2 & -1/2 & 0 & \ldots & 0 \\
-1/2 & 0 & 1/2 & 0 & \ldots & 0 \\
0 & -1/2 & 0 & 1/2 & \ldots & 0 \\
\vdots & \vdots & \ddots & \ddots & \ddots & \vdots \\
0 & \ldots & 0 & 1/2 & -2 & 3/2
\end{bmatrix}_{n \times n}
\]
\subsection{Loss function}
\label{subsec:Loss_function:}

In this study, three different loss functions are utilised for each of the two numerical validation scenario and the experimental validation of the model.
To incorporate physical constraints, temporal derivatives $\dot{z}(t) = {x_t(t), \dot{x}_t(t), g_t(t)}$ are estimated using a tensor differentiator \cite{zhang2019deep}, which implements finite-difference approximations over a local stencil. This component enables embedding of physics-based residuals into the loss function.
For the first scenario of the numerical validation where full-state measurements are available for training, the total loss function used for training combines data loss and physics-based loss:

\begin{equation}
J(\theta)= w_1 J_D(\theta) + w_2 J_P(\theta)
\label{main_eqn}
\end{equation}

\begin{equation}
J_D(\theta) = \frac{1}{N}\sum_{i=1}^{N}\|x^{p(i)}-x^{m(i)}\|_2^2 
+ \frac{1}{N}\sum_{i=1}^{N}\|\dot{x}^{p(i)}-\dot{x}^{m(i)}\|_2^2 
+ \frac{1}{N}\sum_{i=1}^{N}\|g^{p(i)}-g^{m(i)}\|_2^2
\label{eqn_case1_data}
\end{equation}
\begin{equation}    
J_P(\theta) = \frac{1}{N} \sum_{i=1}^{N}\left\| \dot{x}^{p(i)} - x_t^{p(i)} \right\|_2^2 + \frac{1}{N} \sum_{i=1}^{N}\left\| \dot{x}_t^{p(i)} + g^{p(i)} + \Gamma \ddot{x}^{(i)}_g \right\|_2^2
\label{eqn_case1_physics}
\end{equation}

\
Where $J_D(\theta)$ is the data loss based on the measurements and $J_P(\theta)$   denotes the physics loss.
The superscripts $p$ and $m$ denote the predicted and measured (actual) responses, respectively. The index $i$ represents the $i^{\text{th}}$ earthquake sample, and $N$ is the total number of earthquake records used for training. The terms $x$, $\dot{x}$, and $g$ correspond to the structural displacement, velocity, and normalized restoring force, respectively.
 When certain measurement channels are unavailable, corresponding loss terms are omitted. In the physics loss the second term incorporates the physical laws in terms of the governing equation where $\dot{x}_t^p$ represents the derivative of the predicted velocity  (which corresponds to the acceleration).  The first term serves as a consistency or sanity check, ensuring that the predicted state-space variables align with their physical definitions. Specifically, the predicted velocity $\dot{x}^p$ should be equal with the derivative of the predicted displacement($x_t^p$) computed using the tensor differentiator. In this study, equal weights ($w_1 = w_2 = 1$) are used, but these can be tuned depending on the application.
In the other case, although the system information—such as the mass matrix, stiffness, and damping coefficients—is available, only acceleration measurements are used as training data. A modified loss function is architected to incorporate the governing equations and enable training using acceleration data alone to accommodate this limitation. The formulation of this loss function is presented below: 
\begin{equation}
J_P(\theta) = \frac{1}{N} \sum_{i=1}^{N} \left\| \dot{x}^{p(i)} - x_t^{p(i)} \right\|_2^2 +  \frac{1}{N}\sum_{i=1}^{N} \left\| \dot{x}_t^{p(i)} + g^{p(i)} + \Gamma \ddot{x}^{(i)}_g \right\|_2^2,
\label{eqn_case2}
\end{equation}
and 
\begin{equation}
J_D(\theta) =\frac{1}{N} \sum_{i=1}^{N} \left\| \dot{x}_t^{p(i)} - \ddot{x}^{m(i)} \right\|_2^2
\end{equation}
 In the data loss, the $\dot{x}_t^p$ represents the time derivative of the predicted velocity, computed using the tensor differentiator, and enforced to match the available measured acceleration data.

For the experimental validation where we don't have a complete description of the structural system—including mass, stiffness, damping properties, and interaction effect, it is not possible to implement a physics-based loss term, and the model relies solely on the data loss. In this purely data-driven scenario, only acceleration data gathered from the sensors are available for training. The model is designed to predict displacements, which are then differentiated twice using the tensor differentiator to obtain the corresponding accelerations ($x_{tt}$) using the tensor differentiator. The loss function implemented for this case is presented below:
\begin{equation}
J_D(\theta) = \frac{1}{N} \sum_{i=1}^{N} \left\|x_{tt}^{p(i)}   - \ddot{x}^{m(i)} \right\|_2^2
\label{eqn_exp_val}
\end{equation}

\section{Application in Response Prediction of a nonlinear single-degree-of-freedom system:}
\label{subsec: Num Val of the model:}
For numerical validation, we consider a nonlinear single-degree-of-freedom system subjected to ground motion excitation. The system is governed by
\begin{equation}
m \ddot{x} + \underbrace{c \dot{x} + k_1 x + k_2 x^3}_{\text{h}} = -m \Gamma \ddot{x}_g
\label{eq:num1}
\end{equation}
where \( m = 1\,\mathrm{kg} \), \( c = 1\,\mathrm{Ns/m} \), \( k_1 = 20\,\mathrm{N/m} \), and \( k_2 = 200\,\mathrm{N/m} \). The restoring force \( h \) is normalized by mass to yield \( g = h/m \), following Zhang~\cite{phycnn}.  Two scenarios are considered. In the first, full-state measurements—displacement, velocity, and restoring force—are assumed available for training. These can be extracted from numerical simulations, with \( g \) computed via \( g = -\ddot{x} - \Gamma \ddot{x}_g \). In the second, only acceleration data are used, representing a more constrained condition. This setup enables direct assessment of the proposed model’s robustness under varying levels of data availability. The benchmark dataset used for training and validation was obtained from the publicly available PhyCNN repository: https://github.com/zhry10/PhyCNN.

\subsection{\texorpdfstring{Case 1: Full-state measurements available for training}{Case 1: Full-state measurements available for training}}
\label{subsubsec: Num Val of the model1:}
We consider the scenario in which full-state measurements---displacement ($x$), velocity ($\dot{x}$), and restoring force ($g$)---are available for training the model. Following Zhang \cite{phycnn}, we adopt a dataset of 100 numerical simulations of the nonlinear single-degree-of-freedom system\ref{eq:num1} subjected to synthetic earthquake ground motions drawn from the PEER Strong Motion Database~\cite{chiou2008nga}, representing a 10\% probability of exceedance in 50 years. Each simulation spans 50 seconds with a sampling rate of 20 Hz, resulting in 1001 time steps per example. Out of these, 10 records are randomly selected for training, while the remaining 90 are reserved for validation. The proposed PhyULSTM framework (Figure~\ref {Fig1}) is trained to map ground motion acceleration to system response using these data. The model input and output tensors are of shapes [10, 1001, 1] and [10, 1001, 3], respectively. Training follows the architecture and loss function detailed in \ref{subsec:Loss_function:}. 
\[
J(\theta) = w_1 \cdot J_D(\theta) + w_2 \cdot J_P(\theta)
\]
\[
J_D(\theta) =
\frac{1}{N}\sum_{i=1}^{N}\|x^{p(i)}-x^{m(i)}\|_2^2
+ \frac{1}{N}\sum_{i=1}^{N}\|\dot{x}^{p(i)}-\dot{x}^{m(i)}\|_2^2
+ \frac{1}{N}\sum_{i=1}^{N}\|g^{p(i)}-g^{m(i)}\|_2^2
\]
\[
J_P(\theta)
= \frac{1}{N} \sum_{i=1}^{N} \left\| \dot{x}^{p(i)} - x_{t}^{p(i)} \right\|_2^2
+ \frac{1}{N} \sum_{i=1}^{N} \left\| \dot{x}_{t}^{p(i)} + g^{p(i)} + \Gamma \ddot{x}_g^{(i)} \right\|_2^2 
\]

\begin{figure*}
\centering
\includegraphics[width=0.8\textwidth]{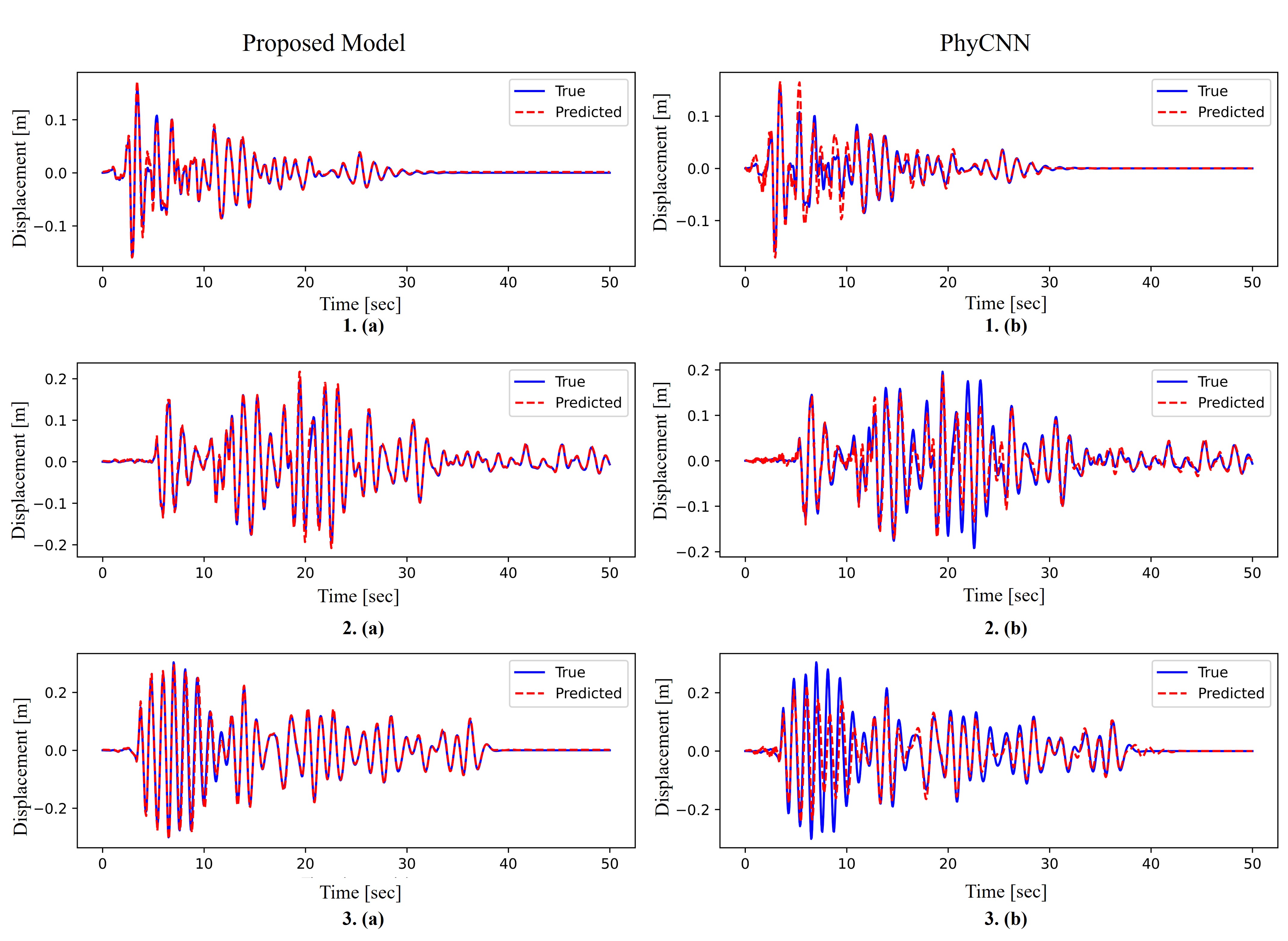}
\caption{\textbf{} Time histories of displacement responses predicted by PhyULSTM (red, left column) and PhyCNN (red, right column) compared against the reference solution (blue). Each row corresponds to a different earthquake input (1, 2, and 3). Subplots are labeled 1a, 1b, 2a, 2b, 3a, and 3b for reference, where the left subplots (a) show PhyULSTM predictions and the right subplots (b) show PhyCNN predictions. }
\label{fig:Numerical_val1}
\end{figure*}

\begin{figure*}
\centering
\includegraphics[width=0.8\textwidth]{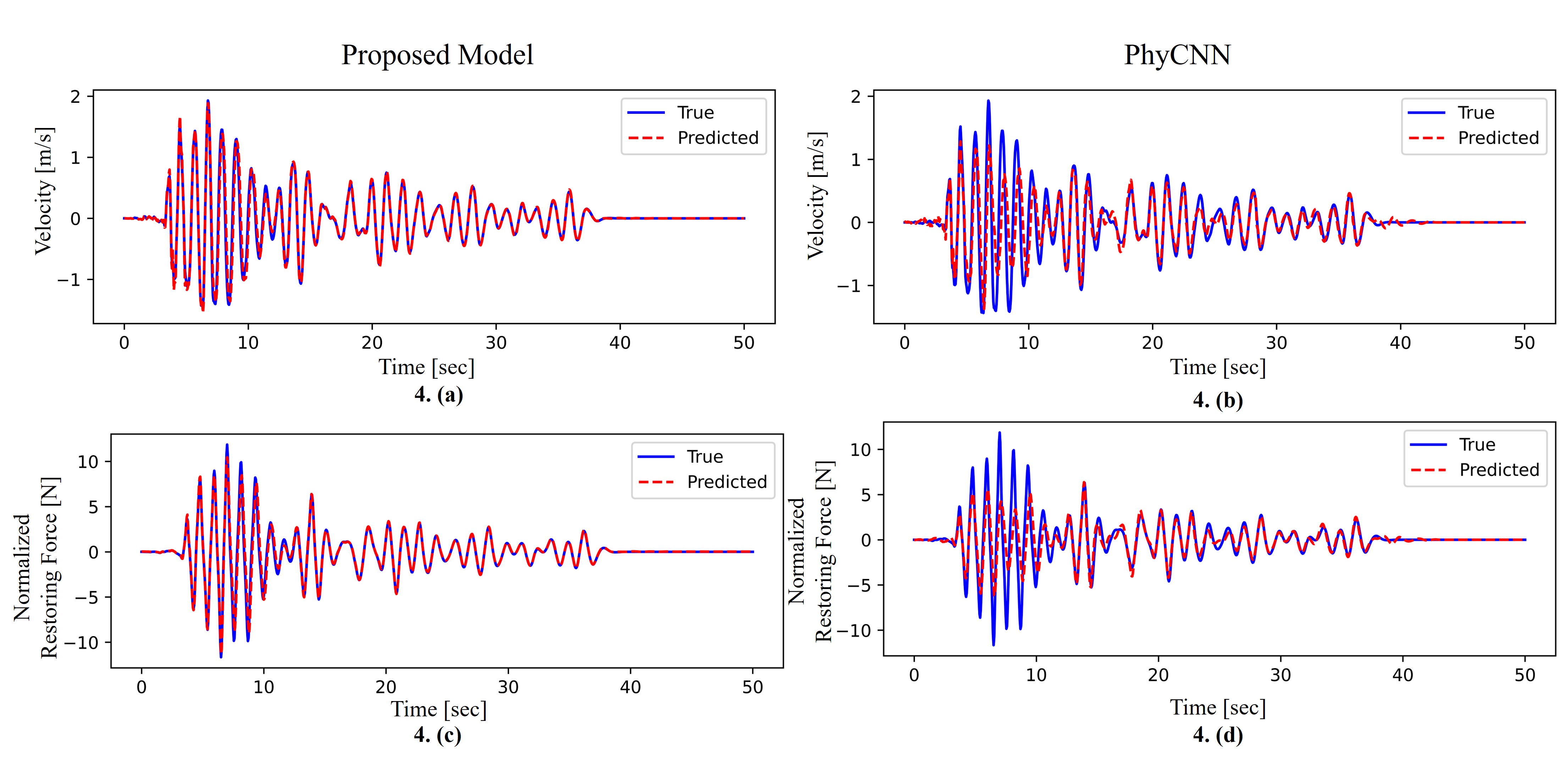}
\caption{\textbf{} Predicted time histories of velocity ($\dot{x}$) and nonlinear restoring force ($g$) for a representative ground motion are shown.The left column shows the responses predicted by PhyULSTM, and the right column shows those predicted by PhyCNN. Subplots are labeled 4a, 4b, 4c, and 4d for reference: 4a and 4c correspond to velocity and normalized restoring force predicted by PhyULSTM, while 4b and 4d correspond to the same predicted by PhyCNN. }
\label{fig:Numerical_val2utg}
\end{figure*}

To evaluate the generalization capability of the proposed framework under unseen seismic excitations, the model predictions are compared against those obtained from the trained PhyCNN architecture. Figure~\ref{fig:Numerical_val1} illustrates the predicted displacement response for three unseen ground motion using both PhyULSTM and PhyCNN models. The proposed PhyULSTM architecture demonstrates excellent predictive performance, accurately reproducing the displacement time history with close agreement to the ground truth. Notably, PhyULSTM captures both the overall trend and the rapid fluctuations in the response that are significantly underrepresented by PhyCNN. This difference is especially prominent in regions with sharp nonlinear behavior, underscoring PhyULSTM’s superior ability to generalize across previously unseen inputs. Figure~\ref{fig:Numerical_val2utg} further compares the predicted velocity and nonlinear restoring force for the a representative ground motion. Again, PhyULSTM closely matches the true values obtained from direct numerical integration of the governing equations. In contrast, PhyCNN tends to smooth out higher-frequency variations and fails to capture abrupt transitions in the restoring force dynamics. This limitation becomes particularly evident in the high-gradient regions, where PhyULSTM maintains fidelity to the true signal.         

\begin{figure*}[!t]
\centering
\begin{subfigure}[t]{0.4\textwidth}
    \centering
    \includegraphics[width=\textwidth]{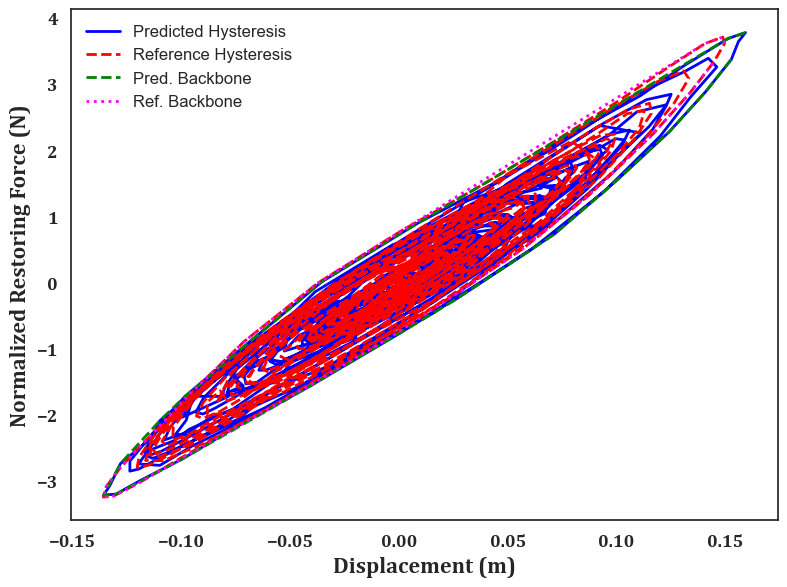}
\end{subfigure}
\begin{subfigure}[t]{0.4\textwidth}
    \centering
    \includegraphics[width=\textwidth]{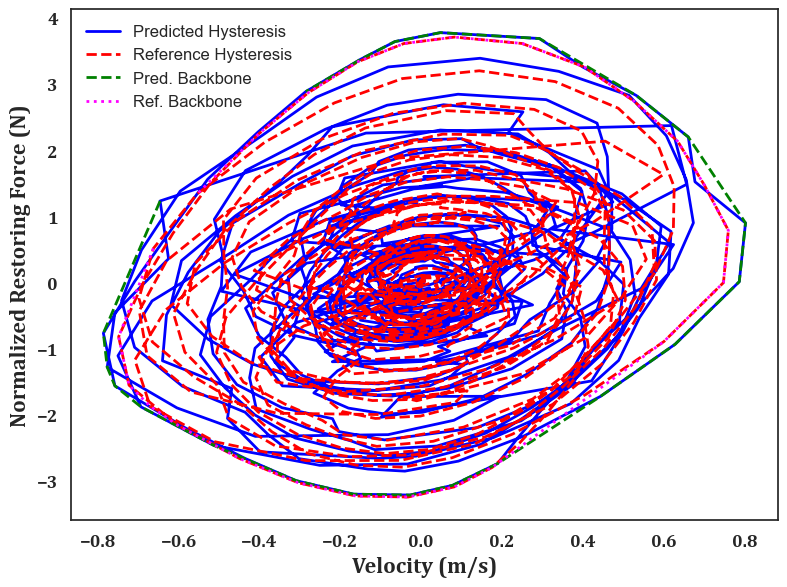}
\end{subfigure}
\caption{\textbf{} PhyULSTM accurately replicates the mechanical response of the nonlinear system, as evidenced by the hysteresis plots of restoring force $g$ versus displacement $x$ (left) and velocity $\dot{x}$ (right). The model successfully captures the characteristic loops, amplitude variations, and shape distortions typical of nonlinear restoring forces, affirming its ability to infer high-order dynamic behavior from limited observations.}
\label{Hyst_numag2u}
\end{figure*}

The nonlinear dynamic behavior of the system is further examined through hysteresis plots of the restoring force versus displacement and velocity, as shown in Figure~\ref{Hyst_numag2u}. These plots serve as critical indicators of the model’s ability to capture complex energy dissipation mechanisms and stiffness degradation inherent in nonlinear structural systems. The results in Figure~\ref{Hyst_numag2u} thus confirm that PhyULSTM is not only capable of accurate time-series predictions but also preserves the nonlinear physics of the underlying system, making it a reliable tool for surrogate modeling in structural dynamics. 

\begin{figure*}
\centering
\includegraphics[width=0.65\textwidth]{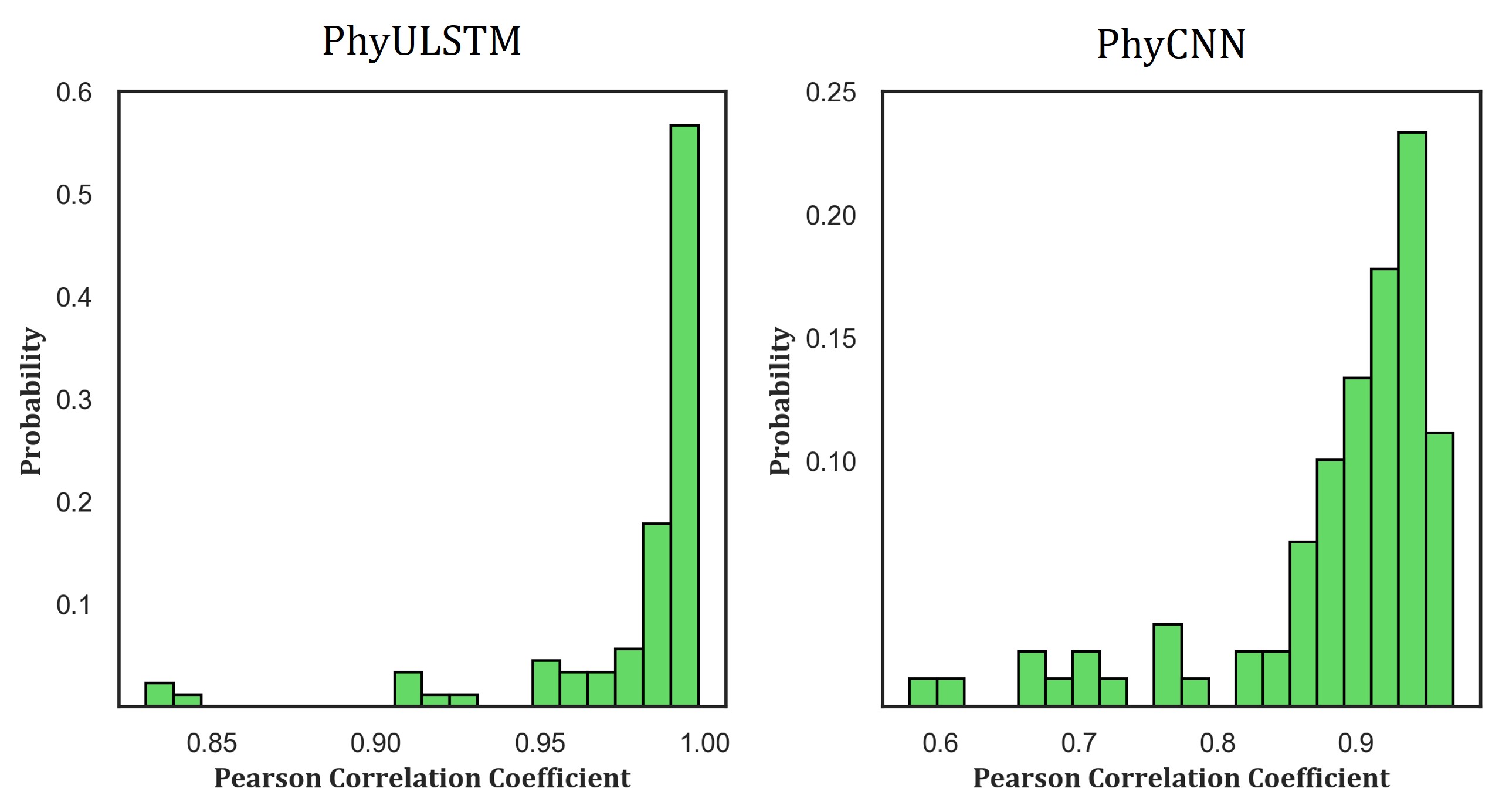}
\caption{\textbf{} Histogram of predicted versus ground-truth displacements for PhyULSTM (left) and PhyCNN (right). PhyULSTM achieves a maximum correlation coefficient of 0.998 and a minimum of 0.83 across all test cases, indicating robust accuracy and generalization. In contrast, PhyCNN exhibits lower performance, with correlation values ranging from 0.97 to 0.577. These results highlight the superior learning capacity of the proposed physics-informed LSTM architecture.}
\label{Reg_numag2u}
\end{figure*}

Figure~\ref{Reg_numag2u} presents the results of a regression analysis comparing the predicted and true displacement responses for both PhyULSTM and PhyCNN models. The proposed PhyULSTM framework demonstrates remarkable predictive fidelity, achieving a maximum correlation coefficient of 0.998 and a minimum of 0.83 across all 90 unseen test cases. These values indicate near-perfect agreement in most scenarios, even under conditions involving significant nonlinearity and variability in input ground motions. In contrast, the PhyCNN model exhibits noticeably lower predictive accuracy, with correlation coefficients ranging from 0.97 down to 0.577. 

Additionally, in the case of the PhyCNN model, 61.1\% of the predicted displacement responses exhibit a Pearson correlation coefficient greater than 0.9 with the reference values. In contrast, the proposed PhyULSTM model achieves a significantly improved performance, with 96.7\% of its predicted displacements exceeding the 0.9 correlation threshold. This comparison underscores the robustness and reliability of PhyULSTM, particularly in scenarios with limited training data. The ability of PhyULSTM to maintain high prediction accuracy across diverse and challenging test inputs highlights its superior generalization capacity and establishes it as a more effective surrogate modeling framework for nonlinear dynamic systems.
\subsubsection{Global error comparison}

The overall prediction accuracy for the test set is summarized in Table~\ref{tab:case1_global_error}. The PhyULSTM model achieves a mean RMSE of $9.52\times10^{-3}$, whereas PhyCNN produces a significantly larger value of $2.46\times10^{-2}$, corresponding to an error reduction of approximately $61\%$. A similar trend is observed for the relative $L_2$ error, where PhyULSTM attains $19.28\%$ compared to $47.40\%$ for PhyCNN. The median values follow the same pattern, confirming that the performance improvement is consistent across the dataset rather than being influenced by a small number of samples. These results demonstrate the superior time-domain prediction accuracy of the proposed PhyULSTM framework.

\begin{table*}[h!]
\centering
\caption{Global error metrics for Case 1 (full-state training).}
\label{tab:case1_global_error}
\begin{tabular}{lcccc}
\hline
Model & Mean RMSE & Median RMSE & Mean Rel-$L_2$ & Median Rel-$L_2$ \\
\hline
PhyULSTM & $9.5160\times10^{-3}$ & $5.6856\times10^{-3}$ & $19.28\%$ & $15.60\%$ \\
PhyCNN   & $2.4625\times10^{-2}$ & $1.6814\times10^{-2}$ & $47.40\%$ & $41.77\%$ \\
\hline
\end{tabular}
\end{table*}

\subsubsection{Frequency-domain comparison}
\begin{figure}[htbp]
\centering
\includegraphics[width=\textwidth]{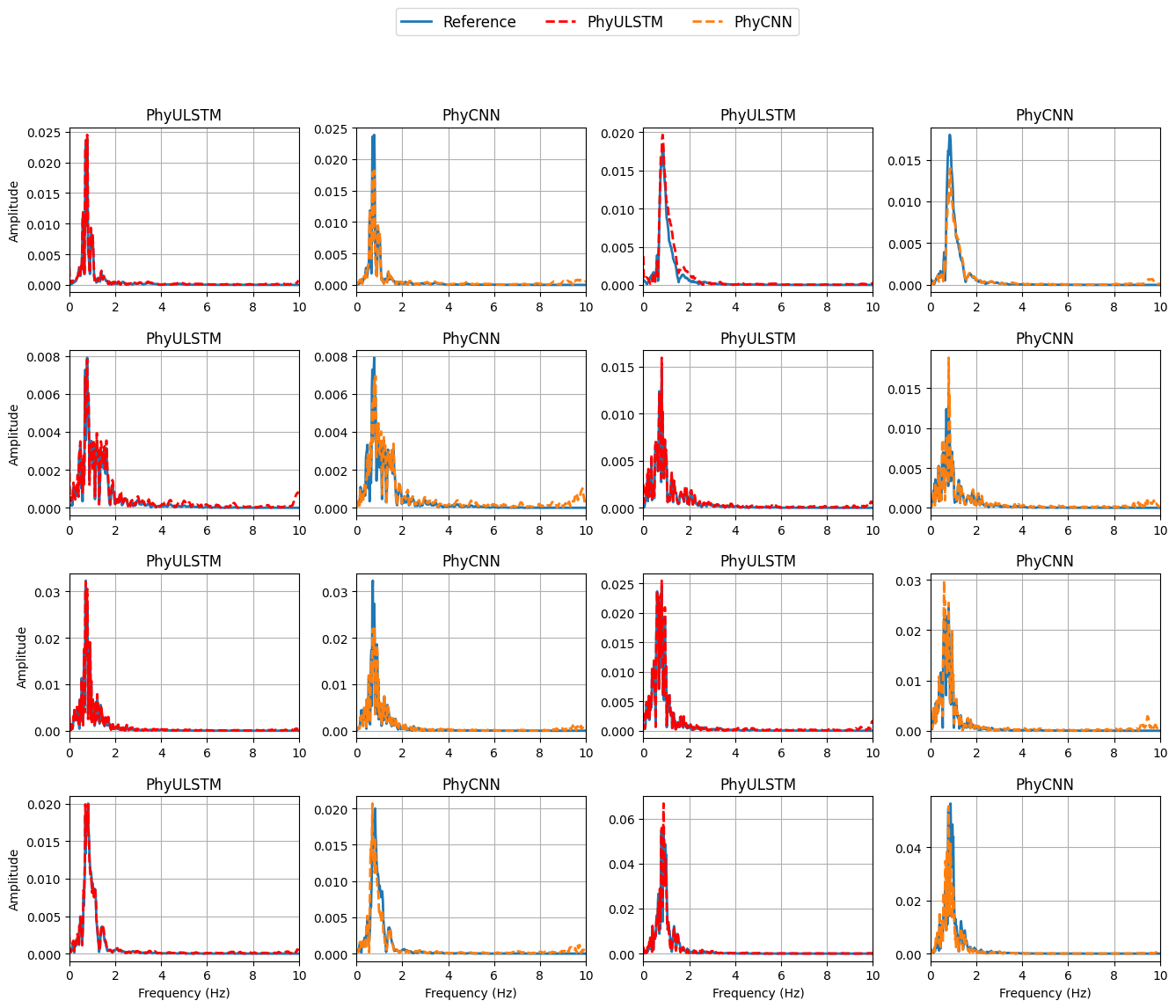}
\caption{Comparison of Fourier amplitude spectra of reference and predicted responses for multiple test samples under full-state training.}
\label{fig:fft_case1}
\end{figure}

To complement the time-domain analysis, Fourier Transform (FFT) analysis was performed on the model predictions and reference responses for 8 random test cases (Fig.~\ref{fig:fft_case1}). The frequency spectra indicate that the system dynamics are dominated by low-frequency components ($< 2$ Hz), characterised by pronounced peaks.

Both PhyULSTM and PhyCNN successfully reproduce the dominant frequency locations and the overall spectral decay, indicating that the primary dynamic characteristics of the structure are captured. However, a closer inspection reveals differences in the prediction of peak spectral amplitudes. In several cases, the PhyCNN model either overestimates or underestimates the peak amplitude, whereas the PhyULSTM predictions consistently show closer agreement with the reference spectra.

Accurate estimation of peak response is vital for reliability and fragility analysis, where extreme response quantities govern structural safety. The improved peak amplitude prediction observed for PhyULSTM therefore provides additional evidence of its superior ability to capture structural dynamic behaviour. This frequency-domain validation complements the time-domain error metrics and further confirms the fidelity of the proposed model.

\subsubsection{Peak response error analysis}

\begin{figure}[h!]
\centering
\includegraphics[width=0.3\textwidth]{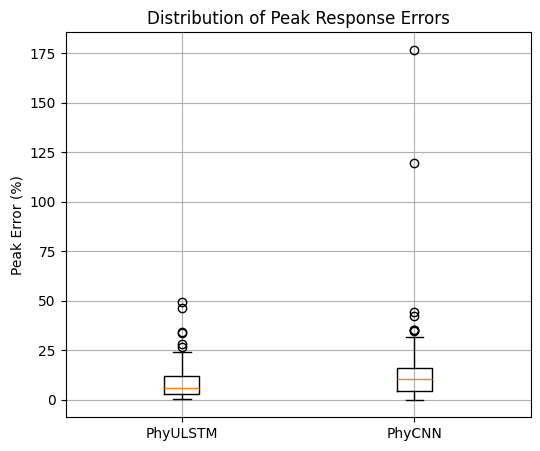}
\caption{Distribution of peak response errors for the test set under full-state training.}
\label{fig:peak_case1}
\end{figure}

The distribution of peak response errors is illustrated in Fig.~\ref{fig:peak_case1}. 
The PhyULSTM model achieves a mean peak error of $8.92\%$ and a median error of $5.76\%$, whereas PhyCNN yields significantly larger values of $14.94\%$ and $10.59\%$, respectively. In addition to the lower central tendency, the PhyULSTM results exhibit a much tighter distribution with fewer extreme outliers. 
In contrast, the PhyCNN predictions show a wider spread and several large outliers, with peak errors exceeding $100\%$ in some test cases. This behavior indicates reduced robustness in predicting extreme structural responses. The improved peak response accuracy of PhyULSTM is particularly important for structural safety and performance assessment, where accurate estimation of maximum responses is critical.

\subsection{\texorpdfstring{Case 2: Training with acceleration measurements only}{Case 2: Training with acceleration measurements only}}
\label{subsubsec: Num Val of the model2:}
\begin{figure*}[ht!]
\centering
\includegraphics[width=0.8\textwidth]{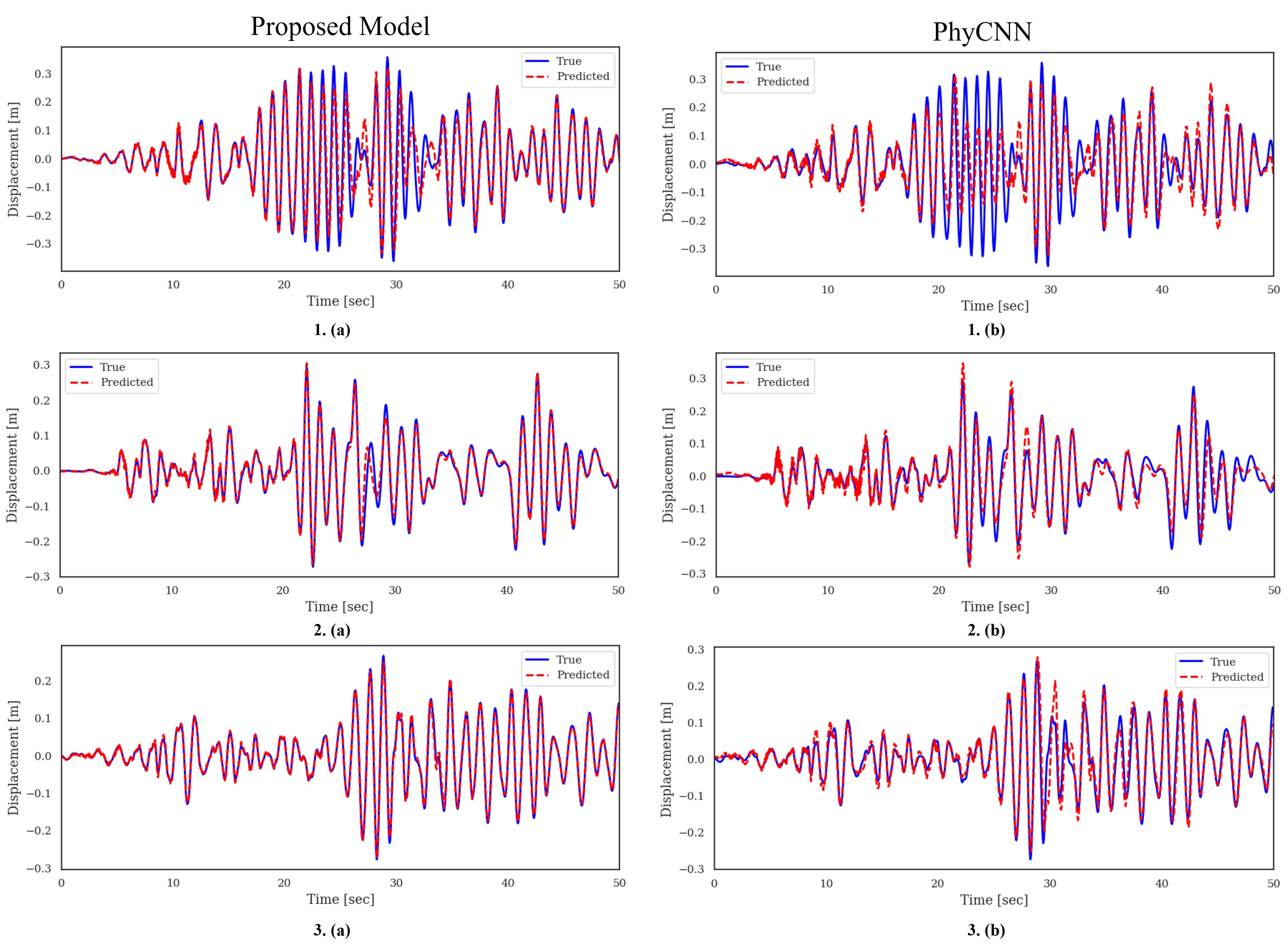}
\caption{\textbf{} Time histories of displacement responses predicted (red) by PhyULSTM (left) and PhyCNN (right) are compared against reference numerical solutions (blue) obtained via direct integration. Each row corresponds to a different earthquake input (1, 2, and 3). Subplots are labeled 1a, 1b, 2a, 2b, 3a, and 3b for reference, where the left subplots (a) show PhyULSTM predictions and the right subplots (b) show PhyCNN predictions. }
\label{fig:Numerical_val2}
\end{figure*}

In practical structural monitoring, accelerometers are often the only sensors installed, providing solely acceleration time histories. Traditional deep learning models approaches attempt to infer displacements by numerical integration of accelerations before training predictive models. However, this introduces significant numerical errors that degrade prediction accuracy. In contrast, the proposed PhyULSTM framework directly incorporates physics constraints into its training, enabling accurate displacement field predictions using only acceleration inputs. Leveraging a specialized 1D U-net architecture combined with LSTM units, the model extracts salient features from acceleration signals to infer the full state vector ${z}(t) = \{x(t), \dot{x}(t), g(t)\}$. These predictions are further refined by a tensor differentiator employing central finite differences to compute derivatives $\dot{z}(t) = \{x_t(t), \dot{x}_t(t), g_t(t)\}$. The loss function is designed to penalize deviations between predicted derivatives and measured accelerations, enforcing physical consistency as
\[
J(\theta) = J_D(\theta) +  J_P(\theta)
\] where \[ J_P(\theta) = \frac{1}{N} \sum_{i=1}^{N} \left\| \dot{x}^{p(i)} - x_t^{p(i)} \right\|_2^2 +  \frac{1}{N}\sum_{i=1}^{N} \left\| \dot{x}_t^{p(i)} + g^{p(i)} + \Gamma \ddot{x}^{(i)}_g \right\|_2^2  \] and \[ J_D(\theta) =\frac{1}{N} \sum_{i=1}^{N} \left\| \dot{x}_t^{p(i)} - \ddot{x}^{m(i)} \right\|_2^2\]
Only acceleration measurements $\ddot{x}^m$ contribute to the loss due to data availability. The model is trained on 50 randomly selected datasets and tested on 50 unseen datasets.

\begin{figure*}[!t]
\centering
\includegraphics[width=0.8\textwidth]{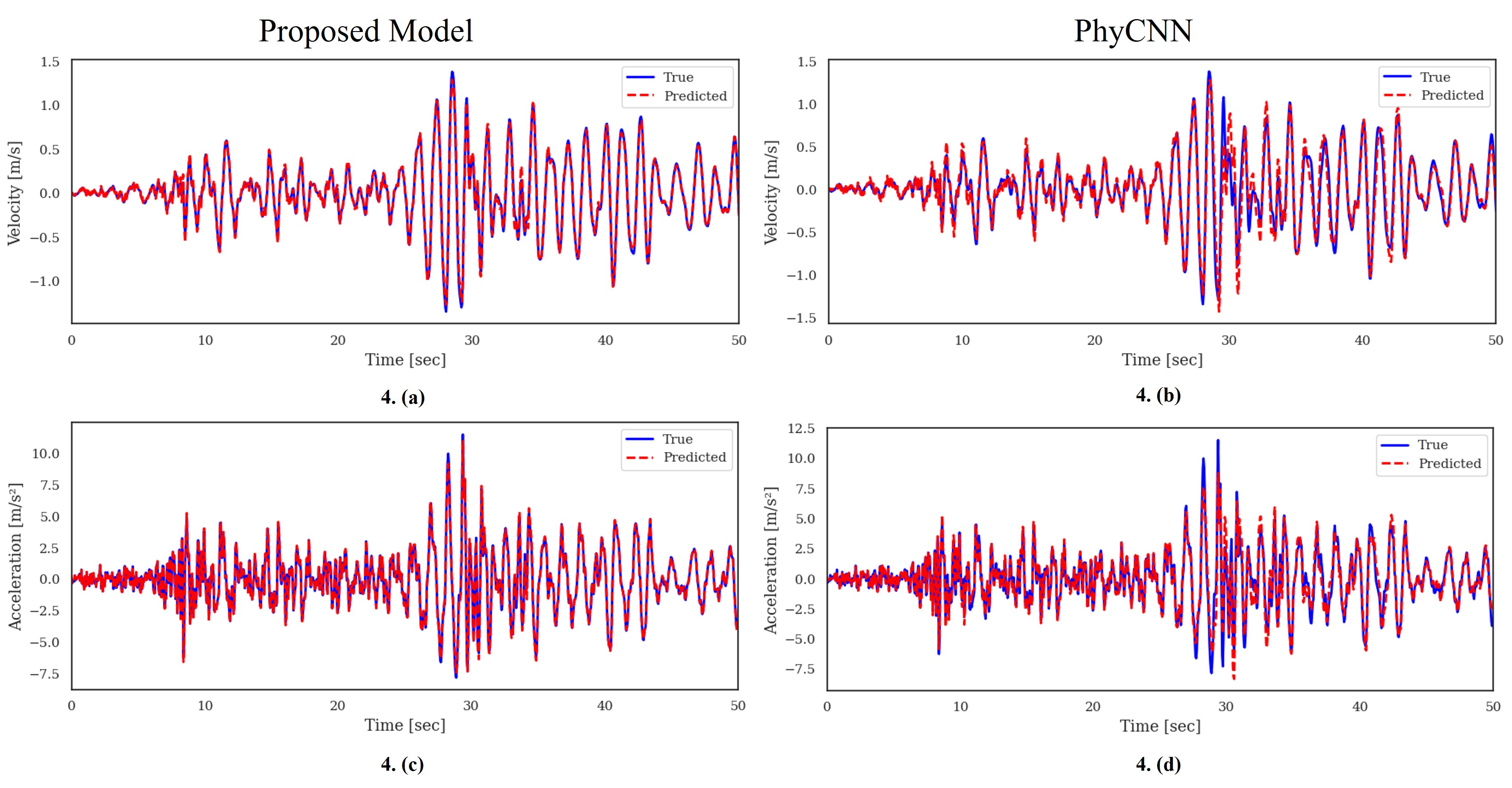}
\caption{\textbf{}PhyULSTM accurately reproduces velocity ($\dot{x}$) and acceleration ($\ddot{x}$) time histories, closely matching the ground truth obtained from numerical integration. The left column shows the responses predicted by PhyULSTM, and the right column shows those predicted by PhyCNN. Subplots are labeled 4a, 4b, 4c, and 4d for reference: 4a and 4c correspond to velocity and acceleration predicted by PhyULSTM, while 4b and 4d correspond to the same quantities predicted by PhyCNN.}
\label{fig:Numerical_val22utg}
\end{figure*}

Figure~\ref{fig:Numerical_val2} illustrates the displacement time histories predicted by the proposed PhyULSTM model alongside those from the baseline PhyCNN and the ground truth numerical solutions for three distinct, previously unseen seismic excitation inputs. The PhyULSTM predictions (left) closely track the true system response (blue), accurately capturing both the amplitude and phase of complex oscillatory behavior inherent to nonlinear structural dynamics. In contrast, the PhyCNN predictions (right) exhibit noticeable deviations, particularly during peak displacement intervals and rapid transient phases, indicating an underestimation of key dynamic features. Figure~\ref{fig:Numerical_val22utg} further highlights the predictive fidelity of PhyULSTM by comparing the velocity ($\dot{x}$) and acceleration ($\ddot{x}$) time histories for a representative unseen seismic excitation. The close agreement between the predicted and reference responses demonstrates the model's capability to accurately capture the underlying nonlinear dynamic behavior of the system.

\begin{figure*}[h]
\centering
\includegraphics[width=0.65\textwidth]{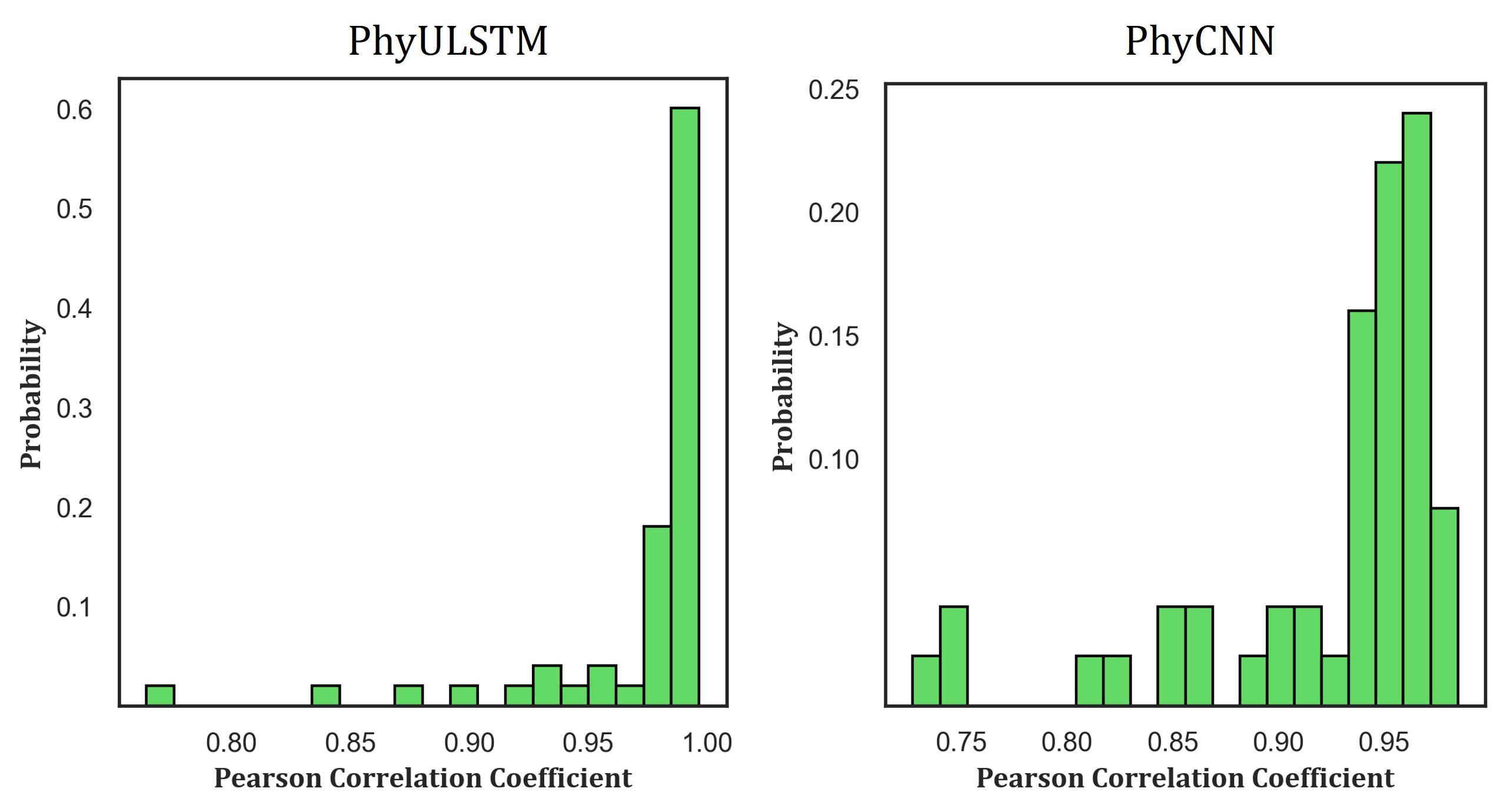}
\caption{\textbf{} Correlation between predicted and true displacements across all test cases is higher for PhyULSTM (left), with maximum and minimum correlation coefficients of 0.996 and 0.764 respectively, compared to PhyCNN (right) which attains 0.984 and 0.727. This demonstrates the superior generalization and predictive robustness of PhyULSTM when trained solely on acceleration data.}
\label{fig:Regressionfornumag2utt}
\end{figure*}
The regression analysis summarised in Figure~\ref{fig:Regressionfornumag2utt} quantitatively confirms these observations. PhyULSTM achieves a maximum correlation coefficient of 0.996 and maintains a minimum of 0.764 across all test cases, demonstrating robust generalization even when trained exclusively with acceleration measurements. By contrast, PhyCNN’s correlation coefficients span a lower range (0.984 maximum to 0.727 minimum), underscoring its comparatively weaker predictive accuracy and generalization. Additionally, in the case of the PhyCNN model, 80\% of the predicted displacement responses exhibit a Pearson correlation coefficient greater than 0.9 with the reference values. In contrast, the proposed PhyULSTM model achieves a significantly improved performance, with 94\% of its predicted displacements exceeding the 0.9 correlation threshold.
 These comprehensive results establish that PhyULSTM is a powerful tool for real-world structural health monitoring and seismic response analysis.
\subsubsection{Global error comparison}

Table~\ref{tab:case2_global_error} summarizes the prediction errors obtained when only acceleration measurements are available during training. The PhyULSTM model achieves a mean RMSE of $1.02\times10^{-2}$ compared to $2.23\times10^{-2}$ for PhyCNN, corresponding to an error reduction of approximately $54\%$. The improvement is even more pronounced in the relative $L_2$ error, where PhyULSTM achieves $20.95\%$ compared to $51.50\%$ for PhyCNN. The median errors follow the same trend, confirming the robustness of the improvement across all test cases.

\begin{table*}[h!]
\centering
\caption{Global error metrics for Case 2 (acceleration-only training).}
\label{tab:case2_global_error}
\begin{tabular}{lcccc}
\hline
Model & Mean RMSE & Median RMSE & Mean Rel-$L_2$ & Median Rel-$L_2$ \\
\hline
PhyULSTM & $1.0195\times10^{-2}$ & $6.7850\times10^{-3}$ & $20.95\%$ & $14.90\%$ \\
PhyCNN   & $2.2338\times10^{-2}$ & $1.6906\times10^{-2}$ & $51.50\%$ & $34.04\%$ \\
\hline
\end{tabular}
\end{table*}

\subsubsection{Frequency-domain comparison}

\begin{figure}[htbp]
\centering
\includegraphics[width=\textwidth]{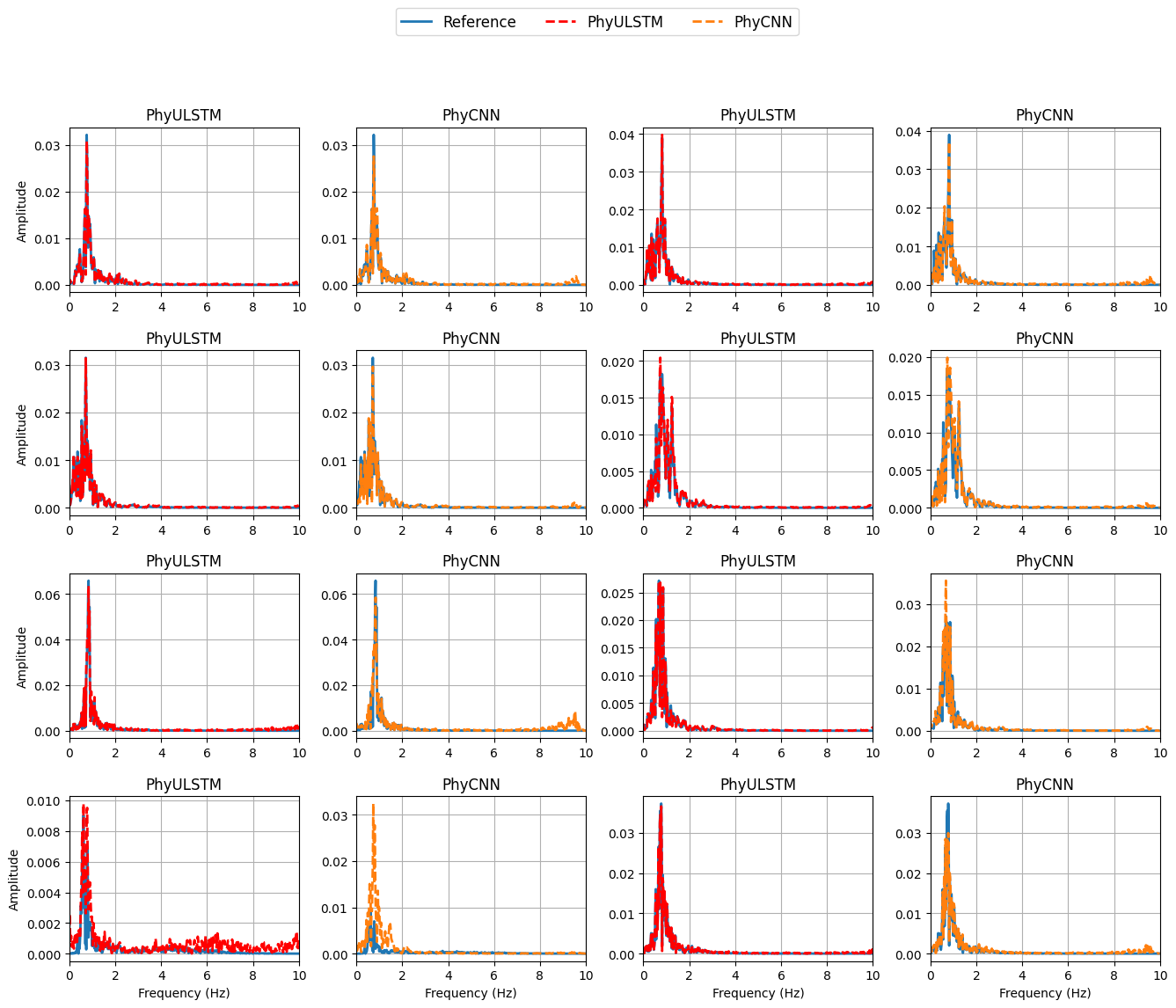}
\caption{Comparison of Fourier amplitude spectra of reference and predicted responses for multiple test samples under acceleration-only training.}
\label{fig:fft_case2}
\end{figure}

The Fourier amplitude spectra of the predicted responses are shown in Fig.~\ref{fig:fft_case2}. The spectra confirm that the structural response remains dominated by low-frequency components ($< 2$ Hz). Both models capture the dominant frequency locations and general spectral decay. 
However, noticeable differences appear in the peak spectral amplitudes. In several cases, the PhyCNN model either overestimates or underestimates the peak amplitude and shows mild spectral spreading at higher frequencies. In contrast, the PhyULSTM predictions consistently show closer agreement with the reference spectra and preserve the dominant peak magnitudes more accurately. This behavior indicates improved reconstruction of structural dynamics when temporal dependencies are modeled.

\subsubsection{Peak response error analysis}

\begin{figure}[h!]
\centering
\includegraphics[width=0.3\textwidth]{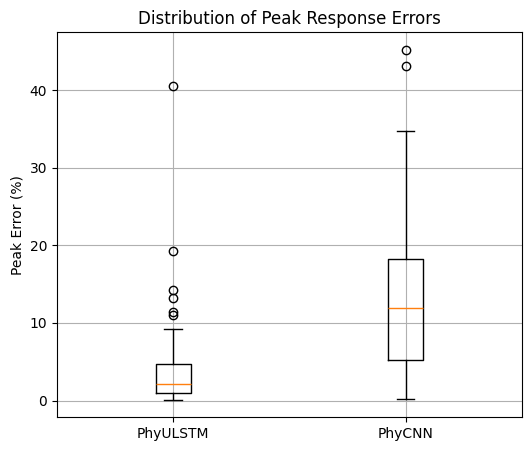}
\caption{Distribution of peak response errors for acceleration-only training.}
\label{fig:peak_case2}
\end{figure}

The distribution of peak response errors is presented in Fig.~\ref{fig:peak_case2}. 
The PhyULSTM model achieves a mean peak error of $4.37\%$ and a median error of $2.17\%$, 
while PhyCNN produces significantly larger values of $12.83\%$ and $11.88\%$, respectively.
In addition to the lower central tendency, the PhyULSTM predictions exhibit a much tighter 
error distribution and substantially fewer extreme outliers.

\FloatBarrier

\section{Application in Response Prediction of a Bouc--Wen Model with Rate-Dependent Hysteresis}
\label{sec:boucwen_model}
The proposed PhyULSTM framework is applied to the response prediction of a nonlinear single-degree-of-freedom (SDOF) system governed by a rate-dependent Bouc--Wen hysteresis model \cite{phylstm}. This benchmark problem is widely used to assess the capability of data-driven and physics-informed models in capturing nonlinear, history-dependent structural behavior.

The equation of motion of the SDOF system subjected to ground excitation is expressed as
\begin{equation}
    m\,\ddot{u}(t) + c\,\dot{u}(t) + \lambda k\,u(t) + (1-\lambda)k\,r(t)
    = -m\,a_g(t)
    \label{eq:boucwen_sdof}
\end{equation}
where $u(t)$ denotes the displacement response, $\dot{u}(t)$ and $\ddot{u}(t)$ are the velocity and acceleration, respectively, $m$, $c$, and $k$ are the mass, damping, and stiffness coefficients, $r(t)$ is the hysteretic internal variable, $\lambda \in (0,1]$ is the post-yield stiffness ratio, and $a_g(t)$ is the ground acceleration.

Normalizing Eq.~\eqref{eq:boucwen_sdof} by the mass $m$ yields
\begin{equation}
    \ddot{u}(t) + 2\zeta\omega_n\,\dot{u}(t) + \lambda\omega_n^2\,u(t) + (1-\lambda)\omega_n^2\,r(t)
    = -a_g(t)
    \label{eq:boucwen_normalized}
\end{equation}
where $\omega_n = \sqrt{k/m}$ is the natural frequency and $\zeta = c/(2\sqrt{km})$ is the damping ratio. The corresponding mass-normalized restoring force is defined as
\begin{equation}
    g(t) = 2\zeta\omega_n\,\dot{u}(t) + \lambda\omega_n^2\,u(t) + (1-\lambda)\omega_n^2\,r(t),
    \label{eq:restoring_force}
\end{equation}
which leads to the compact equation of motion
\begin{equation}
    \ddot{u}(t) + g(t) = -a_g(t).
    \label{eq:compact_eom}
\end{equation}

The evolution of the hysteretic variable $r(t)$ follows the rate-dependent Bouc--Wen differential equation
\begin{equation}
    \dot{r}(t)
    = \dot{u}(t)
    - \alpha |\dot{u}(t)| |r(t)|^{n-1} r(t)
    - \beta \dot{u}(t) |r(t)|^n
    \label{eq:boucwen_hysteresis}
\end{equation}
where $\alpha$ controls the size and shape of the hysteresis loop, $\beta$ governs the smoothness of the transition between elastic and inelastic regimes, and $n \geq 1$ determines the sharpness of yielding. The initial condition is set as $r(0)=0$.
\begin{table}[h!]
\centering
\caption{Bouc--Wen System Parameters}
\label{tab:boucwen_parameters}
\begin{tabular}{lcc}
\toprule
\textbf{Parameter} & \textbf{Symbol} & \textbf{Value} \\
\midrule
Mass & $m$ & 500 kg \\
Damping coefficient & $c$ & 0.35 kN·s/m \\
Stiffness & $k$ & 25 kN/m \\
Bouc--Wen parameter & $\alpha$ & 2.0 \\
Bouc--Wen parameter & $\beta$ & 2.0 \\
Bouc--Wen exponent & $n$ & 3.0 \\
Post-yield stiffness ratio & $\lambda$ & 0.5 \\
\bottomrule
\end{tabular}
\end{table}

\subsection*{System Parameters}
The Bouc--Wen system parameters adopted in this study are summarized in Table~\ref{tab:boucwen_parameters}. The
 corresponding derived system properties are the natural frequency $\omega_n = 7.0711$ rad/s, natural period $T_n = 0.8886$ s, 
 damping ratio $\zeta = 0.007$, and an approximate yield displacement of $u_y \approx 0.01$ m.

\subsection*{Loss Function Formulation}
In this case, the PhyULSTM architecture directly predicts velocity and mass-normalized restoring force, while displacement is
obtained via numerical time integration and acceleration is computed using finite-difference differentiation denoted by  \( {\int}\dot{u} \) and \(\dot{u}_{t}\) respectively. Model training is guided by a composite loss function that combines data fidelity with physics-based constraints:

\begin{equation}
    \mathcal{L}_{\text{total}} = 
    \mathcal{L}_{\text{data}}
    + \mathcal{L}_{\text{physics}}
    \label{eq:total_loss}
\end{equation}
where each term contributes to a distinct aspect of model supervision.

The data loss penalizes discrepancies between predicted and reference structural responses across displacement, velocity, and 
acceleration. For a batch of \(N\) ground motion samples, the data loss is defined as
\begin{multline}\label{eq:data_loss}
\mathcal{L}_{\text{data}} = \frac{1}{N}\sum_{i=1}^{N}
\left\| \int\dot{u}^{p(i)} - u^{true(i)} \right\|_2^2 
+ \frac{1}{N}\sum_{i=1}^{N}
\left\| \dot{u}^{p(i)} - \dot{u}^{true(i)} \right\|_2^2 
+ \frac{1}{N}\sum_{i=1}^{N}
\left\|  \dot{u}_{t}^{p(i)}-\ddot{u}^{true(i)} \right\|_2^2 
\end{multline}

The physics loss enforces dynamic equilibrium. Here, the superscript \( p \) denotes the predicted response. Using $\Phi$ to denote the finite-difference operator, the equilibrium residual for the $i$-th sample is defined as
\[
    r^{(i)} = \dot{u}_{t}^{p(i)} + g^{(i)}_{\text{pred}} + a_g^{(i)}
\]
with $\dot{u}_t = \Phi \dot{u}$. The physics loss is given by
\begin{equation}
    \mathcal{L}_{\text{physics}} =
    \frac{1}{N_c}\sum_{i=1}^{N_c} \| r^{(i)} \|_2^2
    \label{eq:physics_loss}
\end{equation}
where $N$ is the number of training samples and $N_c$ denotes the number of collocation points. Optional weighting parameters may be introduced to balance the relative importance of data fidelity and physics enforcement.Here, the superscript \( p \) denotes the predicted response.

If desired, the different components may be balanced using scalar weights $\lambda_{\text{data}}$, $\lambda_{\text{phys}}$. The weighted total loss reads:
\begin{equation}
    \mathcal{L}_{\text{total}} \;=\; \lambda_{\text{data}}\mathcal{L}_{\text{data}}
    \;+\; \lambda_{\text{phys}}\mathcal{L}_{\text{physics}}
    \label{eq:weighted_total_loss}
\end{equation}
Choosing the weights allows control over the relative importance of data fidelity, and physics enforcement during training.

The system is discretized using a time step $\Delta t = 0.02$ s (sampling frequency $50$ Hz). The Bouc--Wen hysteresis equation~\eqref{eq:boucwen_hysteresis} is integrated using a forward Euler scheme. A synthetic dataset consisting of 100 independent seismic response simulations is generated by exciting the SDOF system with band-limited white noise (BLWN) ground motions of varying intensities. Each simulation spans 30 s, resulting in 1501 time steps per record. Ten BLWN input--response pairs are randomly selected as ``known'' datasets for training and validation using an 80/20 split, while the remaining samples are treated as ``unknown'' datasets for evaluating generalization. In addition, 50 collocation samples containing only ground motion records are included during training to strengthen physics enforcement. The dataset is publicly available at \url{https://github.com/zhry10/PhyLSTM}.
\begin{figure*}[htbp]
    \centering
    \includegraphics[width=0.5\textwidth]{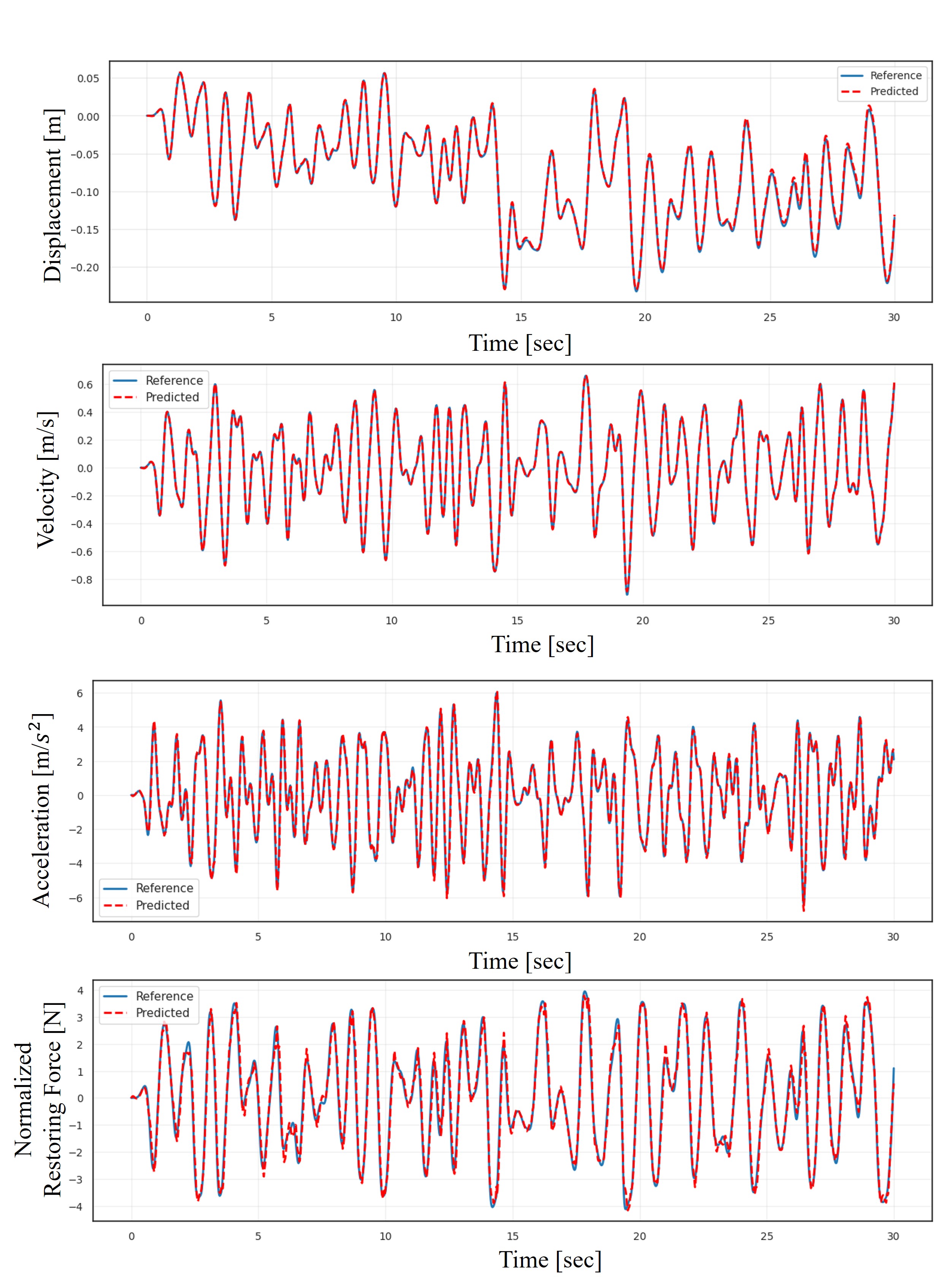}
    \caption{Predicted vs reference responses — displacement, velocity, acceleration, and normalized restoring force for an unseen earthquake from the test dataset.}
    \label{fig:unseen_eq_all_responses}
\end{figure*}
\FloatBarrier
Unlike the PhyLSTM2 and PhyLSTM3 frameworks \cite{phylstm}, which employ multiple LSTM networks to separately model state variables and hysteresis evolution, the proposed PhyULSTM uses a single deep LSTM coupled with a 1D U-Net feature extractor. The network is trained for 5,000 epochs using the Adam optimizer with a learning rate of $10^{-3}$, followed by an additional 5,000 epochs with a reduced learning rate of $10^{-4}$. This unified architecture enforces internal consistency among state variables, mitigates drift accumulation, and improves computational efficiency.

\begin{figure*}
    \centering
    \includegraphics[width=0.85\textwidth]{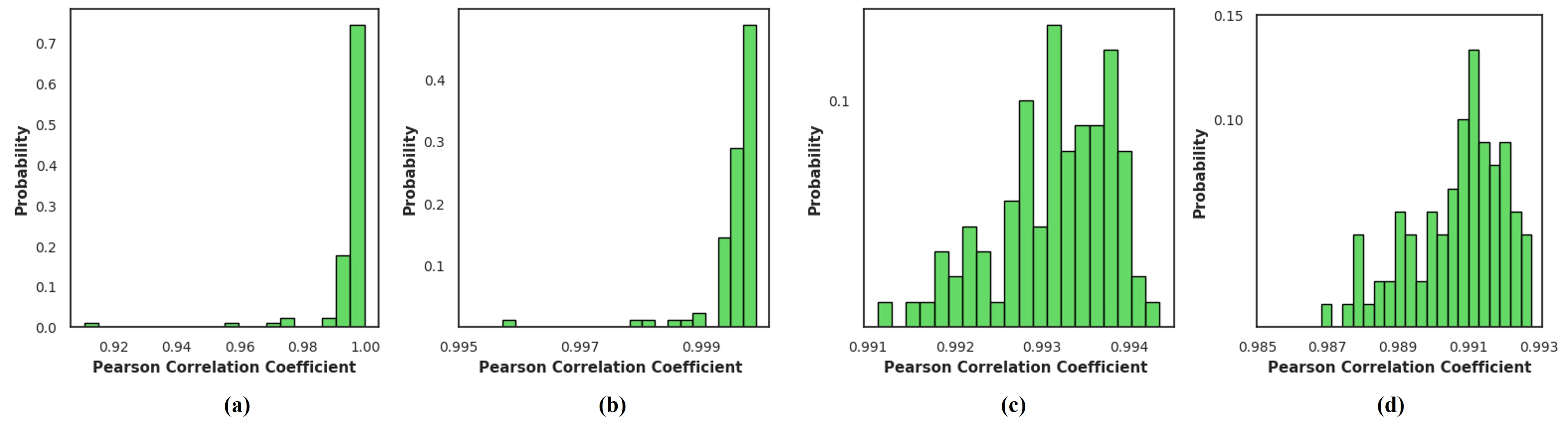}
    \caption{Correlation coefficient versus probability distributions for the predicted displacement (a), velocity (b), acceleration (c), and normalised restoring force (d) responses corresponding to unseen earthquakes from the test dataset.}
    \label{fig:corr_probability_unseen_eq}
\end{figure*}

Figure~\ref{fig:unseen_eq_all_responses} presents displacement, velocity, acceleration, and normalized restoring force responses for an unseen earthquake record.
The PhyULSTM predictions closely match the numerical reference across all response quantities.
Quantitative performance comparisons further highlight the superiority of PhyULSTM. For displacement prediction, correlation coefficients range from $0.911$ to $1.000$, compared to $0.77$--$0.99$ for PhyLSTM3 and $0.19$--$0.85$ for PhyLSTM2. For velocity, acceleration, and restoring force, PhyULSTM consistently achieves correlation coefficients exceeding $0.99$. The probability distributions of correlation coefficients for unseen earthquakes are shown in Figure~\ref{fig:corr_probability_unseen_eq}.

A key distinction between PhyULSTM and PhyLSTM3 lies in the treatment of hysteresis. While PhyLSTM3 explicitly models the hysteretic variable $r(t)$ using a dedicated LSTM network, PhyULSTM learns hysteretic behavior implicitly through its hierarchical feature extraction and temporal memory. These results indicate that, when appropriately regularized by physics-based constraints, deep neural networks can learn physically meaningful, history-dependent representations without explicitly encoding the governing constitutive equations.
\FloatBarrier

\section{Application in response prediction of a six-story reinforced concrete building}
\label{subsec: Exp Val of the model:}
Structural systems are inherently complex, often characterized by nonlinear behavior, heterogeneous materials, and uncertain boundary conditions. In many real-world scenarios, obtaining a complete description of the structural system—including mass, stiffness, damping properties, and interaction effects—is infeasible due to limited instrumentation, inaccessible subsystems, or cost constraints. Traditional numerical methods, while powerful, rely on prior knowledge of the governing equations and boundary conditions. As such, they are generally limited to idealized models and may exhibit significant deviations from actual system behavior when applied to partially known or ill-defined systems. To overcome these limitations, we extend the PhyULSTM framework—originally formulated with embedded physics-based constraints—to operate under data-driven settings where physical knowledge of the system is incomplete or entirely unavailable. In this formulation, the loss function excludes physics-based residual terms, relying solely on measurement-based error minimization. The network architecture remains unchanged, retaining the hybrid encoder–decoder configuration that combines a one-dimensional U-Net with a Long Short-Term Memory (LSTM) decoder . This architecture is capable of learning complex mappings from input ground motions to structural responses, even in the absence of explicit physical regularization.

To validate this data-driven variant of the PhyULSTM model, we utilize recorded seismic response data from a six-story reinforced concrete hotel building located in San Bernardino, California. The structure, designed in 1970, has been instrumented with nine accelerometers on the first, third, and roof levels in both principal directions. The dataset, curated by the Center for Engineering Strong Motion Data (CESMD) \cite{haddadi2012report}, comprises earthquake recordings collected between 1987 and 2018. These events span a broad range of magnitudes and frequency content, thereby providing a representative and challenging benchmark for data-driven structural response modeling. For consistency and reproducibility, the dataset partitioning follows the protocol established in Zhang et al. \cite{phycnn}, which involves stratified sampling into training, validation, and test subsets. The training objective for the data-driven PhyULSTM model is to learn a mapping from ground acceleration inputs to structural displacement outputs. 
\begin{figure}[htbp]
\centering

\begin{minipage}{0.48\textwidth}
\centering
\includegraphics[width=\textwidth]{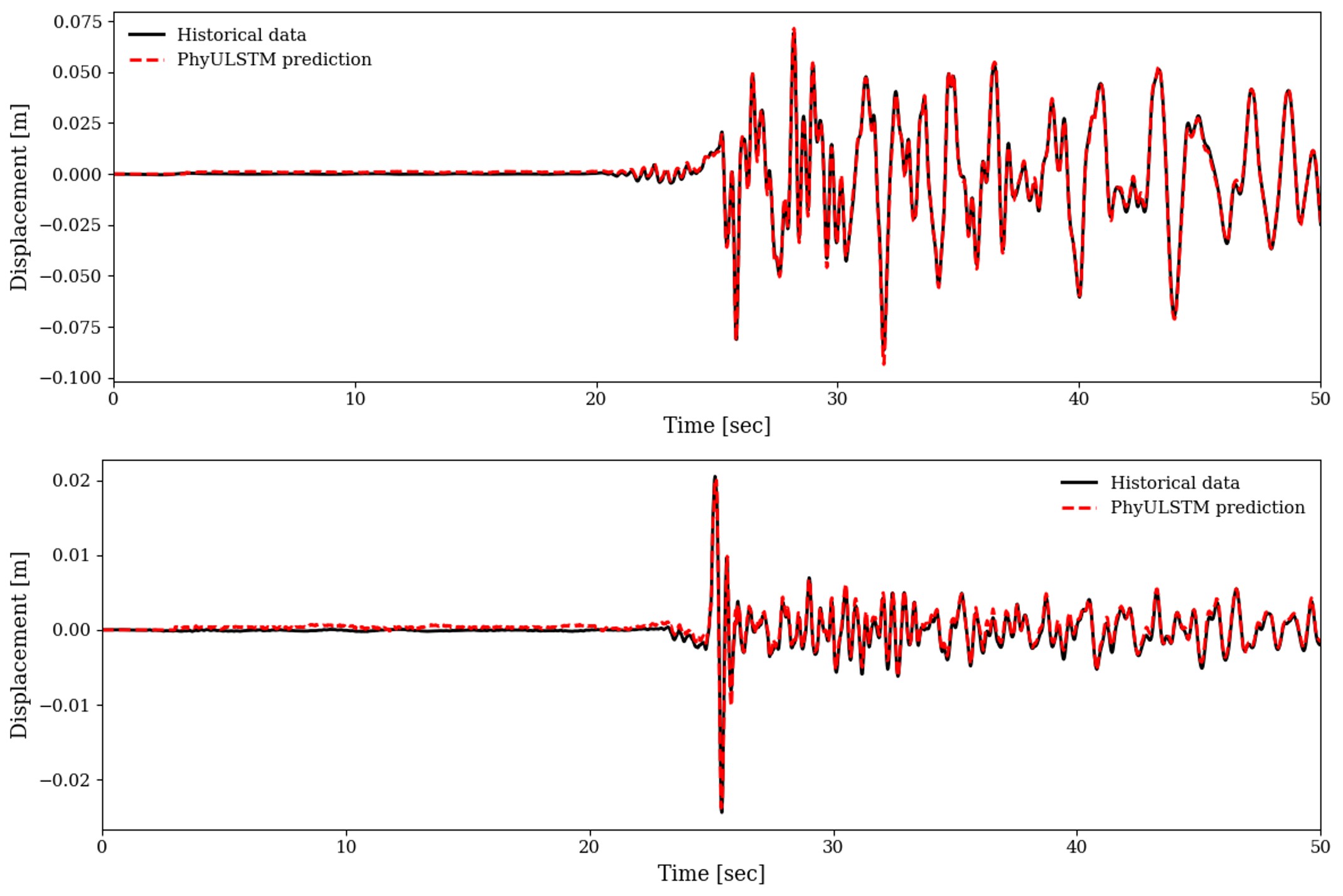}
\caption{Predicted displacement response at the 3rd floor (DOF-1) of the six-story hotel building in San Bernardino, California, subjected to the 2014 Big Bear Lake and 2016 Loma Linda earthquakes.}
\label{fig:expag2utt1}
\end{minipage}
\hfill
\begin{minipage}{0.48\textwidth}
\centering
\includegraphics[width=\textwidth]{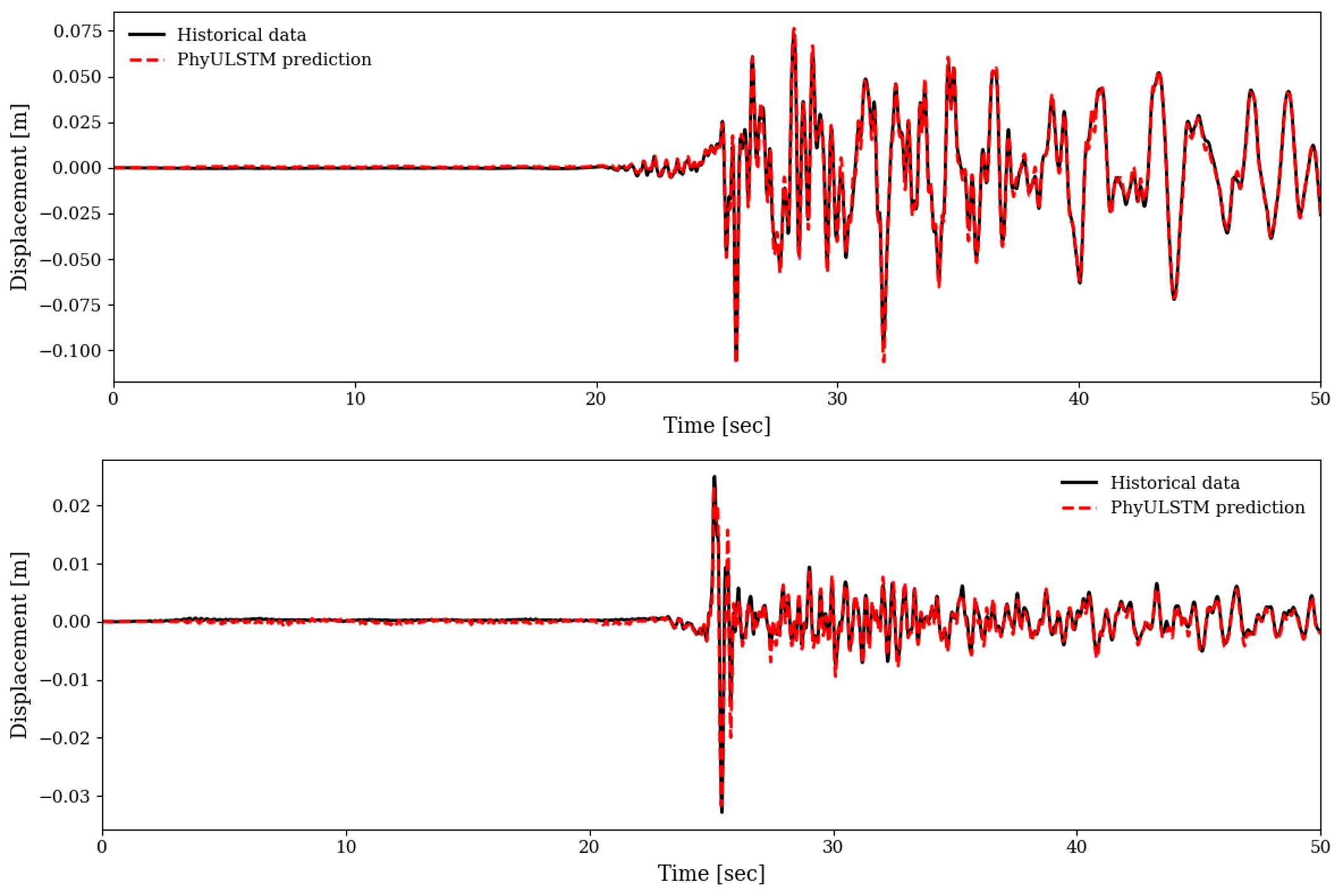}
\caption{Predicted roof-level displacement (DOF-2) of the six-story San Bernardino hotel building under the 2014 Big Bear Lake and 2016 Loma Linda earthquake events.}
\label{fig:expag2utt2}
\end{minipage}

\end{figure}

The model is trained by minimizing the discrepancy between predicted and measured accelerations, defined by the following objective function:

\[
J_D(\theta) = \frac{1}{N} \sum_{i=1}^{N} \left\| \ddot{x}^{m(i)} - x_{tt}^{p(i)} \right\|_2^2
\]

where $\ddot{x}^m$ denotes the measured structural acceleration and $x_{tt}^p$ represents the predicted acceleration obtained by applying a central finite difference operator to the predicted displacement time series. The network is trained using 11 earthquake records, with 4 events used for validation. Once trained, the model is tested on previously unseen events to evaluate its generalization capability. Figure~\ref{fig:expag2utt1} and Figure~\ref{fig:expag2utt2} present the predicted displacement responses at the 3rd floor and roof levels, respectively, of the six-story instrumented hotel building in San Bernardino, California, subjected to the 2014 Big Bear Lake and 2016 Loma Linda earthquakes. The PhyULSTM model was trained using only acceleration time histories obtained from sensors distributed across the lower, middle, and upper stories of the structure, without incorporating explicit physics-based constraints in the loss function. Despite this, the model accurately captures essential structural behaviors across events of varying intensity, duration, and frequency content. 
These findings highlight the robustness and generalization capacity of the PhyULSTM architecture for modeling real-world nonlinear structural systems under seismic excitation. Importantly, the model achieves this without requiring detailed system identification or constitutive modeling, positioning it as a viable surrogate modeling tool for structural health monitoring applications, particularly in scenarios where only partial or noisy sensor data are available.

\subsection{Error Distribution Comparison}

\begin{figure}[htbp]
\centering
\includegraphics[width=0.5\textwidth]{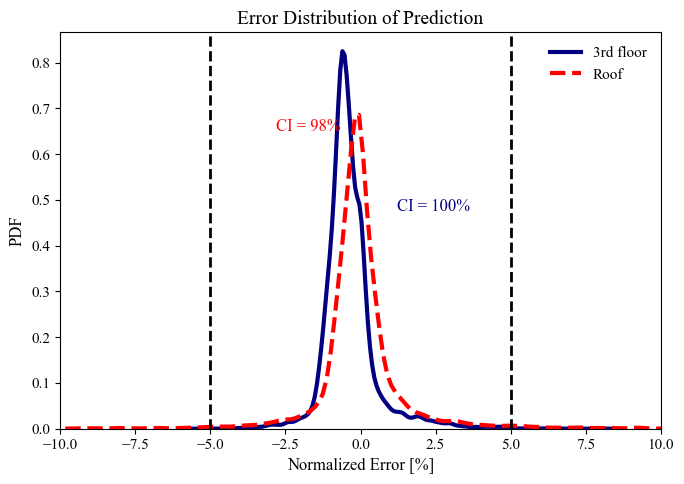}
\caption{Probability density function of normalised prediction error for the 3rd floor and roof displacement responses. The dashed vertical lines indicate the $\pm5\%$ error bounds. The proposed model achieves confidence intervals of 99.89\% and 98.05\% for the 3rd floor and roof, respectively.}
\label{fig:error_distribution}
\end{figure}

To evaluate the prediction accuracy of the proposed framework, the probability density function (PDF) of the normalised displacement error is analysed and compared with the results reported for the PhyCNN model. The normalised error is defined as the difference between the predicted and reference responses divided by the maximum absolute reference displacement.
\begin{equation}
e = \frac{y_{\text{true}} - y_{\text{pred}}}{\max(|y_{\text{true}}|)},
\end{equation}
where $y_{\text{true}}$ and $y_{\text{pred}}$ denote the measured and predicted displacements, respectively. The probability density function (PDF) of this normalized error is estimated via kernel density estimation and plotted for both the 3rd floor and roof levels in Figure\ref{fig:error_distribution}.
\subsubsection{Confidence Interval Comparison}
Comparison of prediction error confidence intervals within $\pm 5\%$ is presented in Table \ref{tab:conf interval}. For the PhyCNN model, the error distribution shows that the prediction error is predominantly concentrated within $\pm 5\%$, with confidence intervals (CI) of 97\% for the 3rd floor and 93\% for the roof. The PDFs are sharply peaked near zero with minimal tails beyond the $\pm 5\%$ bounds, demonstrating high point-wise accuracy across the prediction set.
In comparison, the proposed PhyULSTM achieves comparable or slightly superior performance on the same benchmark case (six-story hotel building, acceleration-only inputs, and identical data partitioning via K-means clustering). Specifically, the CI within the $\pm 5\%$ error band is 99.89\% for the 3rd floor and 98.05\% for the roof (exceeding PhyCNN's 93\%). The error PDFs for both levels remain tightly concentrated around zero, with lower dispersion at the roof level, as evidenced by the higher CI and narrower peak in the plot. This indicates excellent agreement with the reference responses, particularly for the upper stories where long-range temporal dependencies are more pronounced.

\begin{table}[h]
\centering
\begin{tabular}{c c c}
\hline
Model & 3rd Floor & Roof \\
\hline
PhyCNN & 97\% & 93\% \\
 PhyULSTM& \textbf{99.89\%} & \textbf{98.05\%} \\
\hline
\end{tabular}
\caption{Comparison of prediction error confidence intervals within $\pm 5\%$.}
\label{tab:conf interval}
\end{table}
\FloatBarrier

\subsubsection{Global error comparison}
Both models achieve very small absolute errors, indicating that the overall structural response is well captured. 

PhyULSTM achieves a mean RMSE of $1.13\times10^{-3}$ compared to $1.44\times10^{-3}$ for PhyCNN, corresponding to an improvement of approximately $22\%$. The relative $L_2$ error is also slightly lower for PhyULSTM ($0.23\%$ vs $0.28\%$), confirming its superior accuracy. Although the difference in global errors is modest due to the high accuracy of both models, the advantage of PhyULSTM becomes more evident when peak responses are examined.

The global error metrics are summarized in Table ~\ref{tab:rc_global_error}.

\begin{table}[h]
\centering
\resizebox{\columnwidth}{!}{%
\begin{tabular}{lcccc}
\hline
Model & Mean RMSE & Median RMSE & Mean Rel-$L_2$ & Median Rel-$L_2$ \\
\hline
PhyULSTM & $1.1289\times10^{-3}$ & $1.0152\times10^{-3}$ & 0.23\% & 0.26\% \\
PhyCNN   & $1.4386\times10^{-3}$ & $1.2763\times10^{-3}$ & 0.28\% & 0.33\% \\
\hline
\end{tabular}
}
\caption{Global error metrics for the six-story RC building case.}
\label{tab:rc_global_error}
\end{table}
\FloatBarrier
\subsubsection{Peak response error analysis}
 The difference between the models becomes significantly more pronounced for peak response prediction. PhyULSTM achieves a mean peak error of $11.88\%$ and a median error of $9.78\%$, whereas PhyCNN exhibits substantially larger errors of $27.65\%$ and $32.76\%$, respectively. This shows that PhyCNN produces larger peak errors and a wider spread, while PhyULSTM maintains a more compact distribution. This result is significant because structural reliability and fragility assessments depend primarily on accurate estimation of extreme responses rather than average behavior.

\FloatBarrier
\section{Conclusion}

This study proposes a novel Physics-Informed U-Net–LSTM (PhyULSTM) framework for the rapid prediction of seismic responses in nonlinear structural systems. This framework effectively addresses key limitations of standalone CNN and LSTM architectures in modeling nonlinear structural dynamics. The proposed framework is benchmarked against established physics-informed architectures, including PhyCNN and PhyLSTM, through two numerical case studies incorporating nonlinear stiffness behavior and a Bouc–Wen hysteretic model, as well as experimental validation using recorded sensor data from a mid-rise reinforced concrete building in San Bernardino. In all scenarios, the PhyULSTM demonstrated superior performance based on correlation coefficients, global error metrics, and peak response prediction accuracy. For cases where full-state measurements were available, 96.7\% of predicted displacement responses achieved correlation coefficients exceeding 0.9, while in the acceleration-only training scenario, 94\% exceeded this threshold. The mean and median peak response prediction errors were 8.92\% and 5.76\%, respectively, for the full-state case, and further reduced to 4.37\% and 2.17\% for the acceleration-only configuration. For the Bouc–Wen nonlinear hysteretic model, displacement prediction correlations ranged from 0.911 to 1.00, while velocity, acceleration, and restoring force predictions consistently closed to 0.99. For the mid-rise reinforced concrete building, the normalized prediction errors are close to zero: 99.89\% of the third-floor predictions and 98.05\% of the roof predictions fall within a ±5\% error bound.

Beyond conventional statistical evaluation, the framework is further assessed for its ability to reproduce nonlinear hysteretic behavior, and for its ability to preserve critical frequency-domain characteristics essential for structural safety and reliability analysis. Through hysteresis loop reconstruction, Fourier amplitude spectrum analysis, and peak response error evaluation, and it is found that the model accurately captures nonlinear response evolution, dominant frequency components, and extreme structural demands. Importantly, the proposed architecture exhibits strong predictive capability under data scarce scenarios. For example, without explicit knowledge of structural parameters such as mass, stiffness, or damping matrices, the framework maintained reliable performance based only on acceleration measurements. This flexibility highlights its robustness for practical deployment in real-time structural health monitoring applications.

Overall, this study establishes PhyULSTM as a reliable, interpretable, and physically grounded metamodeling tool that advances the state of surrogate modeling for structural dynamics and provides a foundation for future developments in hybrid physics-informed machine learning. Future work will focus on enhancing model generalizability across various structural typologies, extending the framework toward inverse problems and parameter identification for structural health monitoring, and improving robustness under broader classes of seismic excitations.

\section*{Code Availability}

The code corresponding to this work will be made publicly available after publication at the following repository:
\url{https://github.com/SuirthaBiswas/PhyULSTM}
\bibliographystyle{unsrt}
\bibliography{references}

@article{stoller2019seq,
  title={Seq-u-net: A one-dimensional causal u-net for efficient sequence modelling},
  author={Stoller, Daniel and Tian, Mi and Ewert, Sebastian and Dixon, Simon},
  journal={arXiv preprint arXiv:1911.06393},
  year={2019}
}

@article{raissi2018deep,
  title={Deep hidden physics models: Deep learning of nonlinear partial differential equations},
  author={Raissi, Maziar},
  journal={Journal of Machine Learning Research},
  volume={19},
  number={25},
  pages={1--24},
  year={2018}
}

@article{raissi2019physics,
  title={Physics-informed neural networks: A deep learning framework for solving forward and inverse problems involving nonlinear partial differential equations},
  author={Raissi, Maziar and Perdikaris, Paris and Karniadakis, George E},
  journal={Journal of Computational physics},
  volume={378},
  pages={686--707},
  year={2019},
  publisher={Elsevier}
}

@article{sun2020surrogate,
  title={Surrogate modeling for fluid flows based on physics-constrained deep learning without simulation data},
  author={Sun, Luning and Gao, Han and Pan, Shaowu and Wang, Jian-Xun},
  journal={Computer Methods in Applied Mechanics and Engineering},
  volume={361},
  pages={112732},
  year={2020},
  publisher={Elsevier}
}

@article{zhu2019physics,
  title={Physics-constrained deep learning for high-dimensional surrogate modeling and uncertainty quantification without labeled data},
  author={Zhu, Yinhao and Zabaras, Nicholas and Koutsourelakis, Phaedon-Stelios and Perdikaris, Paris},
  journal={Journal of Computational Physics},
  volume={394},
  pages={56--81},
  year={2019},
  publisher={Elsevier}
}

@article{phylstm,
  title={Physics-informed multi-LSTM networks for metamodeling of nonlinear structures},
  author={Zhang, Ruiyang and Liu, Yang and Sun, Hao},
  journal={Computer Methods in Applied Mechanics and Engineering},
  volume={369},
  pages={113226},
  year={2020},
  publisher={Elsevier}
}

@article{phycnn,
  title={Physics-guided convolutional neural network (PhyCNN) for data-driven seismic response modeling},
  author={Zhang, Ruiyang and Liu, Yang and Sun, Hao},
  journal={Engineering Structures},
  volume={215},
  pages={110704},
  year={2020},
  publisher={Elsevier}
}

@article{bengio1994learning,
  title={Learning long-term dependencies with gradient descent is difficult},
  author={Bengio, Yoshua and Simard, Patrice and Frasconi, Paolo},
  journal={IEEE transactions on neural networks},
  volume={5},
  number={2},
  pages={157--166},
  year={1994},
  publisher={IEEE}
}

@inproceedings{trinh2018learning,
  title={Learning longer-term dependencies in rnns with auxiliary losses},
  author={Trinh, Trieu and Dai, Andrew and Luong, Thang and Le, Quoc},
  booktitle={International Conference on Machine Learning},
  pages={4965--4974},
  year={2018},
  organization={PMLR}
}

@article{bai2018empirical,
  title={An empirical evaluation of generic convolutional and recurrent networks for sequence modeling},
  author={Bai, Shaojie and Kolter, J Zico and Koltun, Vladlen},
  journal={arXiv preprint arXiv:1803.01271},
  year={2018}
}

@article{oord2016wavenet,
  title={Wavenet: A generative model for raw audio},
  author={Oord, Aaron van den and Dieleman, Sander and Zen, Heiga and Simonyan, Karen and Vinyals, Oriol and Graves, Alex and Kalchbrenner, Nal and Senior, Andrew and Kavukcuoglu, Koray},
  journal={arXiv preprint arXiv:1609.03499},
  year={2016}
}

@article{unet1,
  title={U-net and its variants for medical image segmentation: A review of theory and applications},
  author={Siddique, Nahian and Paheding, Sidike and Elkin, Colin P and Devabhaktuni, Vijay},
  journal={IEEE access},
  volume={9},
  pages={82031--82057},
  year={2021},
  publisher={IEEE}
}

@article{unet2,
  title={Medical Image Segmentation based on U-Net: A Review.},
  author={Du, Getao and Cao, Xu and Liang, Jimin and Chen, Xueli and Zhan, Yonghua},
  journal={Journal of Imaging Science \& Technology},
  volume={64},
  number={2},
  year={2020}
}

@article{wiskott2002slow,
  title={Slow feature analysis: Unsupervised learning of invariances},
  author={Wiskott, Laurenz and Sejnowski, Terrence J},
  journal={Neural computation},
  volume={14},
  number={4},
  pages={715--770},
  year={2002},
  publisher={MIT Press}
}

@inproceedings{ronneberger2015u,
  title={U-net: Convolutional networks for biomedical image segmentation},
  author={Ronneberger, Olaf and Fischer, Philipp and Brox, Thomas},
  booktitle={Medical image computing and computer-assisted intervention--MICCAI 2015: 18th international conference, Munich, Germany, October 5-9, 2015, proceedings, part III 18},
  pages={234--241},
  year={2015},
  organization={Springer}
}

@article{chiou2008nga,
  title={NGA project strong-motion database},
  author={Chiou, Brian and Darragh, Robert and Gregor, Nick and Silva, Walter},
  journal={Earthquake Spectra},
  volume={24},
  number={1},
  pages={23--44},
  year={2008},
  publisher={SAGE Publications Sage UK: London, England}
}

@article{zhang2019deep,
  title={Deep long short-term memory networks for nonlinear structural seismic response prediction},
  author={Zhang, Ruiyang and Chen, Zhao and Chen, Su and Zheng, Jingwei and B{\"u}y{\"u}k{\"o}zt{\"u}rk, Oral and Sun, Hao},
  journal={Computers \& Structures},
  volume={220},
  pages={55--68},
  year={2019},
  publisher={Elsevier}
}

@inproceedings{haddadi2012report,
  title={Report on progress at the center for engineering strong motion data (CESMD)},
  author={Haddadi, H and Shakal, A and Huang, M and Parrish, J and Stephens, C and Savage, W and Leith, W},
  booktitle={The 15th world conference on earthquake engineering. Lisbon, Portugal},
  pages={24--28},
  year={2012}
}

@article{chollet2018keras,
  title={Keras: The python deep learning library},
  author={Chollet, Fran{\c{c}}ois and others},
  journal={Astrophysics source code library},
  pages={ascl--1806},
  year={2018}
}

@article{NN_ref1,
  title={Approximation capabilities of multilayer feedforward networks},
  author={Hornik, Kurt},
  journal={Neural networks},
  volume={4},
  number={2},
  pages={251--257},
  year={1991},
  publisher={Elsevier}
}

@article{NN_ref2,
  title={Approximations of continuous functionals by neural networks with application to dynamic systems},
  author={Chen, Tianping and Chen, Hong},
  journal={IEEE Transactions on Neural networks},
  volume={4},
  number={6},
  pages={910--918},
  year={1993},
  publisher={IEEE}
}

@article{NN_ref3,
  title={Neural networks for nonlinear dynamic system modelling and identification},
  author={Chen, SABS and Billings, Stephen A},
  journal={International journal of control},
  volume={56},
  number={2},
  pages={319--346},
  year={1992},
  publisher={Taylor \& Francis}
}

@article{NN_ref4,
  title={Predicting seismic response of structures by artificial neural networks},
  author={Yu'ao, He and Xianzhong, Hu and Sheng, Zhan},
  journal={Transaction of Tianjin University},
  volume={2},
  number={2},
  pages={36--39},
  year={1996}
}

@article{NN_ref5,
  title={Identification of nonlinear dynamical systems using multilayered neural networks},
  author={Jagannathan, Sarangapani and Lewis, Frank L},
  journal={Automatica},
  volume={32},
  number={12},
  pages={1707--1712},
  year={1996},
  publisher={Elsevier}
}

@article{NN_ref6,
  title={Analysis and modification of Volterra/Wiener neural networks for the adaptive identification of non-linear hysteretic dynamic systems},
  author={Pei, J-S and Smyth, AW and Kosmatopoulos, EB},
  journal={Journal of Sound and Vibration},
  volume={275},
  number={3-5},
  pages={693--718},
  year={2004},
  publisher={Elsevier}
}

@article{NN_ref7,
  title={A neural network approach for structural identification and diagnosis of a building from seismic response data},
  author={Huang, Chiung-Shiann and Hung, Shih-Lin and Wen, CM and Tu, TT},
  journal={Earthquake engineering \& structural dynamics},
  volume={32},
  number={2},
  pages={187--206},
  year={2003},
  publisher={Wiley Online Library}
}

@article{NN_ref8,
  title={Identification of restoring forces in non-linear vibration systems using fuzzy adaptive neural networks},
  author={Liang, YC and Feng, DP and Cooper, JE},
  journal={Journal of sound and vibration},
  volume={242},
  number={1},
  pages={47--58},
  year={2001},
  publisher={Elsevier}
}

@article{NN_ref9,
  title={Neural identification of non-linear dynamic structures},
  author={Le Riche, Rodolphe and Gualandris, David and Thomas, Jean Jacques and Hemez, F},
  journal={Journal of Sound and vibration},
  volume={248},
  number={2},
  pages={247--265},
  year={2001},
  publisher={Elsevier}
}

@article{yinfeng2008nonlinear,
  title={Nonlinear structural response prediction based on support vector machines},
  author={Yinfeng, Dong and Yingmin, Li and Ming, Lai and Mingkui, Xiao},
  journal={Journal of Sound and Vibration},
  volume={311},
  number={3-5},
  pages={886--897},
  year={2008},
  publisher={Elsevier}
}

@article{wu2019deep,
  title={Deep convolutional neural network for structural dynamic response estimation and system identification},
  author={Wu, Rih-Teng and Jahanshahi, Mohammad R},
  journal={Journal of Engineering Mechanics},
  volume={145},
  number={1},
  pages={04018125},
  year={2019},
  publisher={American Society of Civil Engineers}
}

@article{lagaros2012neural,
  title={Neural network based prediction schemes of the non-linear seismic response of 3D buildings},
  author={Lagaros, Nikos D and Papadrakakis, Manolis},
  journal={Advances in Engineering Software},
  volume={44},
  number={1},
  pages={92--115},
  year={2012},
  publisher={Elsevier}
}

@article{graves2012long,
  title={Long short-term memory},
  author={Graves, Alex and Graves, Alex},
  journal={Supervised sequence labelling with recurrent neural networks},
  pages={37--45},
  year={2012},
  publisher={Springer}
}

@article{bird2020cross,
  title={Cross-domain MLP and CNN transfer learning for biological signal processing: EEG and EMG},
  author={Bird, Jordan J and Kobylarz, Jhonatan and Faria, Diego R and Ek{\'a}rt, Anik{\'o} and Ribeiro, Eduardo P},
  journal={IEEE Access},
  volume={8},
  pages={54789--54801},
  year={2020},
  publisher={IEEE}
}

@article{tang2019convolutional,
  title={Convolutional neural network-based data anomaly detection method using multiple information for structural health monitoring},
  author={Tang, Zhiyi and Chen, Zhicheng and Bao, Yuequan and Li, Hui},
  journal={Structural Control and Health Monitoring},
  volume={26},
  number={1},
  pages={e2296},
  year={2019},
  publisher={Wiley Online Library}
}

@inproceedings{danilczyk2021smart,
  title={Smart grid anomaly detection using a deep learning digital twin},
  author={Danilczyk, William and Sun, Yan Lindsay and He, Haibo},
  booktitle={2020 52nd North American Power Symposium (NAPS)},
  pages={1--6},
  year={2021},
  organization={IEEE}
}

@article{zhao2019speech,
  title={Speech emotion recognition using deep 1D \& 2D CNN LSTM networks},
  author={Zhao, Jianfeng and Mao, Xia and Chen, Lijiang},
  journal={Biomedical signal processing and control},
  volume={47},
  pages={312--323},
  year={2019},
  publisher={Elsevier}
}

@article{pantidis2024fenn,
  title={I-FENN with Temporal Convolutional Networks: Expediting the load-history analysis of non-local gradient damage propagation},
  author={Pantidis, Panos and Eldababy, Habiba and Abueidda, Diab and Mobasher, Mostafa E},
  journal={Computer Methods in Applied Mechanics and Engineering},
  volume={425},
  pages={116940},
  year={2024},
  publisher={Elsevier}
}

@article{pantidis2023integrated,
  title={Integrated Finite Element Neural Network (I-FENN) for non-local continuum damage mechanics},
  author={Pantidis, Panos and Mobasher, Mostafa E},
  journal={Computer Methods in Applied Mechanics and Engineering},
  volume={404},
  pages={115766},
  year={2023},
  publisher={Elsevier}
}

@article{huang2021deep,
  title={Deep learning for nonlinear seismic responses prediction of subway station},
  author={Huang, Pengfei and Chen, Zhiyi},
  journal={Engineering Structures},
  volume={244},
  pages={112735},
  year={2021},
  publisher={Elsevier}
}

@inproceedings{li2021dynamic,
  title={Dynamic response prediction of vehicle-bridge interaction system using feedforward neural network and deep long short-term memory network},
  author={Li, Huile and Wang, Tianyu and Wu, Gang},
  booktitle={Structures},
  volume={34},
  pages={2415--2431},
  year={2021},
  organization={Elsevier}
}

@article{li2022new,
  title={A new dam structural response estimation paradigm powered by deep learning and transfer learning techniques},
  author={Li, Yangtao and Bao, Tengfei and Gao, Zhixin and Shu, Xiaosong and Zhang, Kang and Xie, Lunchen and Zhang, Zhentao},
  journal={Structural Health Monitoring},
  volume={21},
  number={3},
  pages={770--787},
  year={2022},
  publisher={SAGE Publications Sage UK: London, England}
}

@article{liu2023pi,
  title={PI-LSTM: Physics-informed long short-term memory network for structural response modeling},
  author={Liu, Fangyu and Li, Junlin and Wang, Linbing},
  journal={Engineering Structures},
  volume={292},
  pages={116500},
  year={2023},
  publisher={Elsevier}
}

@article{segura2020metamodel,
  title={Metamodel-based seismic fragility analysis of concrete gravity dams},
  author={Segura, Rocio and Padgett, Jamie E and Paultre, Patrick},
  journal={Journal of Structural Engineering},
  volume={146},
  number={7},
  pages={04020121},
  year={2020},
  publisher={American Society of Civil Engineers}
}

@article{gharehbaghi2020estimating,
  title={Estimating inelastic seismic response of reinforced concrete frame structures using a wavelet support vector machine and an artificial neural network},
  author={Gharehbaghi, Sadjad and Yazdani, Hessam and Khatibinia, Mohsen},
  journal={Neural Computing and Applications},
  volume={32},
  number={8},
  pages={2975--2988},
  year={2020},
  publisher={Springer}
}

@article{vaidyanathan2005artificial,
  title={Artificial neural networks for predicting the response of structural systems with viscoelastic dampers},
  author={Vaidyanathan, CV and Kamatchi, P and Ravichandran, R},
  journal={Computer-Aided Civil and Infrastructure Engineering},
  volume={20},
  number={4},
  pages={294--302},
  year={2005},
  publisher={Wiley Online Library}
}

@article{kim2020probabilistic,
  title={Probabilistic evaluation of seismic responses using deep learning method},
  author={Kim, Taeyong and Song, Junho and Kwon, Oh-Sung},
  journal={Structural Safety},
  volume={84},
  pages={101913},
  year={2020},
  publisher={Elsevier}
}

@article{kim2019response,
  title={Response prediction of nonlinear hysteretic systems by deep neural networks},
  author={Kim, Taeyong and Kwon, Oh-Sung and Song, Junho},
  journal={Neural Networks},
  volume={111},
  pages={1--10},
  year={2019},
  publisher={Elsevier}
}

@article{lecun1995convolutional,
  title={Convolutional networks for images, speech, and time series},
  author={LeCun, Yann and Bengio, Yoshua and others},
  journal={The handbook of brain theory and neural networks},
  volume={3361},
  number={10},
  pages={1995},
  year={1995},
  publisher={Citeseer}
}

@article{reddy1993introduction,
  title={An introduction to the finite element method},
  author={Reddy, Junuthula Narasimha},
  journal={New York},
  volume={27},
  number={14},
  year={1993}
}

@book{huebner2001finite,
  title={The finite element method for engineers},
  author={Huebner, Kenneth H and Dewhirst, Donald L and Smith, Douglas E and Byrom, Ted G},
  year={2001},
  publisher={John Wiley \& Sons}
}

@book{zienkiewicz2005finite,
  title={The finite element method: its basis and fundamentals},
  author={Zienkiewicz, Olgierd Cecil and Taylor, Robert Leroy and Zhu, Jian Z},
  year={2005},
  publisher={Elsevier}
}

@article{kuo2024gnn,
  title={GNN-LSTM-based fusion model for structural dynamic responses prediction},
  author={Kuo, Po-Chih and Chou, Yuan-Tung and Li, Kuang-Yao and Chang, Wei-Tze and Huang, Yin-Nan and Chen, Chuin-Shan},
  journal={Engineering Structures},
  volume={306},
  pages={117733},
  year={2024},
  publisher={Elsevier}
}

@article{newmark1959method,
  title={A method of computation for structural dynamics},
  author={Newmark, Nathan M},
  journal={Journal of the engineering mechanics division},
  volume={85},
  number={3},
  pages={67--94},
  year={1959},
  publisher={American Society of Civil Engineers}
}

@article{kolay2014development,
  title={Development of a family of unconditionally stable explicit direct integration algorithms with controllable numerical energy dissipation},
  author={Kolay, Chinmoy and Ricles, James M},
  journal={Earthquake Engineering \& Structural Dynamics},
  volume={43},
  number={9},
  pages={1361--1380},
  year={2014},
  publisher={Wiley Online Library}
}

@article{lecun1989handwritten,
  title={Handwritten digit recognition with a back-propagation network},
  author={LeCun, Yann and Boser, Bernhard and Denker, John and Henderson, Donnie and Howard, Richard and Hubbard, Wayne and Jackel, Lawrence},
  journal={Advances in neural information processing systems},
  volume={2},
  year={1989}
}

@article{lecun2002gradient,
  title={Gradient-based learning applied to document recognition},
  author={LeCun, Yann and Bottou, L{\'e}on and Bengio, Yoshua and Haffner, Patrick},
  journal={Proceedings of the IEEE},
  volume={86},
  number={11},
  pages={2278--2324},
  year={2002},
  publisher={Ieee}
}

@article{krizhevsky2012imagenet,
  title={Imagenet classification with deep convolutional neural networks},
  author={Krizhevsky, Alex and Sutskever, Ilya and Hinton, Geoffrey E},
  journal={Advances in neural information processing systems},
  volume={25},
  year={2012}
}

\end{document}